\documentclass[letterpaper, 10 pt, conference]{ieeeconf}
\IEEEoverridecommandlockouts

\usepackage[utf8]{inputenc}
\usepackage[T1]{fontenc}
\usepackage{graphicx}
\usepackage{float} 
\usepackage{ragged2e}
\usepackage{booktabs}
\usepackage{bbding}
\usepackage{multirow}
\usepackage{makecell}
\usepackage[linesnumbered,ruled]{algorithm2e}
\usepackage{amsmath,amssymb,amsfonts}
\usepackage[table,xcdraw]{xcolor}
\usepackage{cite}
\usepackage[colorlinks,linkcolor=blue]{hyperref}
\usepackage{cleveref}
\usepackage{tabularx}
\usepackage[caption=true,font=normalsize,labelfont=sf,textfont=sf]{subfig}
\usepackage{textcomp}
\usepackage{stfloats}

\hypersetup{
    colorlinks=true, 
    linkcolor=blue,  
    filecolor=blue,  
    urlcolor=black,   
    citecolor=blue,  
    pdfborder={0 0 0} 
}

\Crefname{figure}{Fig.}{Figs.}
\Crefname{equation}{Eq.}{Eqs.}

\def\eg{\emph{e.g.}}

\def\ie{\emph{i.e.}}

\newcommand{\verticaltext}[2][0pt]{%
  \raisebox{#1}{\parbox[t]{1em}{\rotatebox[origin=c]{90}{#2}}}%
}

\newtheorem{theorem}{Theorem}

\title{\LARGE \bf
FedEMA: Federated Exponential Moving Averaging with Negative Entropy Regularizer in Autonomous Driving
}

\author{Wei-Bin Kou$^{1,2,3}$, Guangxu Zhu$^{3}$, Bingyang Cheng$^{1}$, Shuai Wang$^{4}$, \\Ming Tang$^{2, *}$, and Yik-Chung Wu$^{1,*}$
\thanks{$^{*}$Corresponding author: Ming Tang (tangm3@sustech.edu.cn) and Yik-Chung Wu (ycwu@eee.hku.hk).}%
\thanks{$^{1}$Department of Electrical and Electronic Engineering, The University of Hong Kong, Hong Kong, China.}%
\thanks{$^{2}$Department of Computer Science and Engineering, Southern University of Science and Technology, Shenzhen, China.}%
\thanks{$^{3}$Shenzhen International Center For Industrial And Applied Mathematics, Shenzhen Research Institute of Big Data, Shenzhen, China.}%
\thanks{$^{4}$Shenzhen Institutes of Advanced Technology, Chinese Academy of Sciences, Shenzhen, China.}%
}

\begin{document}

\maketitle
\thispagestyle{empty}
\pagestyle{empty}

\begin{abstract}
Street Scene Semantic Understanding (denoted as S3U) is a crucial but complex task for autonomous driving (AD) vehicles. Their inference models typically face poor generalization due to domain-shift. Federated Learning (FL) has emerged as a promising paradigm for enhancing the generalization of AD models through privacy-preserving distributed learning. However, these FL AD models face significant \textbf{temporal catastrophic forgetting} when deployed in dynamically evolving environments, where continuous adaptation causes abrupt erosion of historical knowledge. This paper proposes \underline{Fed}erated \underline{E}xponential \underline{M}oving \underline{A}verage (FedEMA), a novel framework that addresses this challenge through two integral innovations: (I) Server-side model's historical fitting capability preservation via fusing current FL round's aggregation model and a proposed previous FL round's exponential moving average (EMA) model; (II) Vehicle-side negative entropy regularization to prevent FL models' possible overfitting to EMA-introduced temporal patterns. Above two strategies empower FedEMA a dual-objective optimization that balances model generalization and adaptability. In addition, we conduct theoretical convergence analysis for the proposed FedEMA. Extensive experiments both on Cityscapes dataset and Camvid dataset demonstrate FedEMA's superiority over existing approaches, showing 7.12\% higher mean Intersection-over-Union (mIoU).
\end{abstract}

\section{INTRODUCTION}
Street Scene Semantic Understanding (termed as S3U) is an essential yet challenging task for autonomous driving (AD) vehicles \cite{10416354,10160999,10049523,kou2025enhancing}. Recent developments in this area have led to a variety of innovative approaches \cite{10342110,10342254,10341639} that have achieved notable successes. Despite these advances, S3U methods commonly encounter generalization challenges due to domain shifts \cite{muhammad2022vision}. Federated Learning (FL) \cite{https://doi.org/10.48550/arxiv.1602.05629,10342134,wu2024hierarchical,kou2024pfedlvm} paradigm has become instrumental in enhancing the generalization of AD model via learning collaboratively while preserving data privacy. However, as AD systems increasingly operate in continuously evolving environments---where weather patterns, traffic regulations, and road infrastructure change over time---a critical flaw emerges: models trained via conventional FL suffer temporal catastrophic forgetting, abruptly losing previously acquired knowledge when adapting to new temporal contexts.

\begin{figure}[!t]
\vspace{-0.4cm}
\centering
\includegraphics[width=\linewidth]{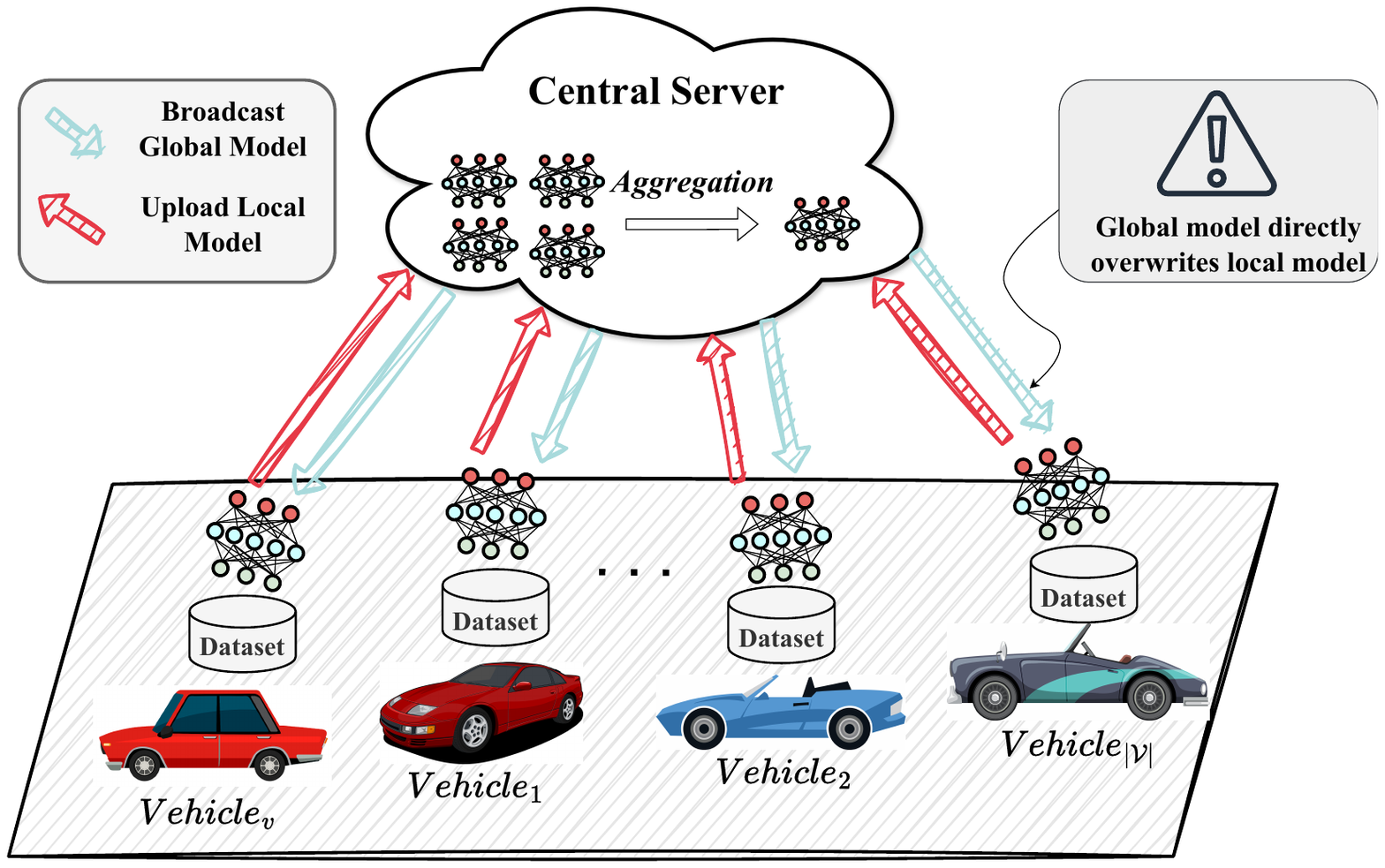}
\vspace{-1.0cm}
\caption{Illustration of Federated AD (FedAD) system \cite{kou2024pfedlvm}. 
The inherent training scheme of FedAD, which continuously overwrites old local model with new global model, inevitably leads to the catastrophic forgetting problem, particularly as the dataset is dynamic over time.}
\label{fig:FedAD_system}
\vspace{-0.55cm}
\end{figure}

This catastrophic forgetting phenomenon stems from FL's overwriting training scheme and not preserving historical model fitting capabilities in lifelong federated AD (FedAD) systems \cite{kou2024fedrc,kou2025fast} (shown in \Cref{fig:FedAD_system}). Specifically, in each training round, the server-side aggregated model is distributed to all involved vehicles to overwrite the local models. This results in that local models just can best fit round-wise data if the data residing vehicles is dynamic over time. Therefore, preserving historical FedAD model is a promising method to mitigate catastrophic forgetting problem. To this end, we should consider three challenges:
\begin{enumerate}
    \item \textbf{How to avoid short-term bias:} Vehicle updates prioritize immediate data, discarding long-term environmental patterns. Thus, how to avoid this biasedness towards short-term scenes is supposed to be handled appropriately. 
    \item \textbf{How to conduct historical model fusion:} Overly rigid preservation of historical parameters stifles adaptation to new scenarios, while excessive focus on recent data erases hard-won foundational knowledge. As a result, how to make a good trade-off between the preservation of historical model and the focus on current model should be necessarily dealt with.
    \item \textbf{How to penalize model complexity to avoid temporal overfitting:} While preserving historical model in FedAD systems, the model is prune to overfitting to temporal patterns. Therefore, how to penalize model complexity to generalize it across wide range of scenarios should be properly considered.
\end{enumerate}

\begin{figure*}[!t]
\vspace{-0.4cm}
\hspace{-0.7cm}
\includegraphics[width=1.08\linewidth, height=0.9\linewidth]{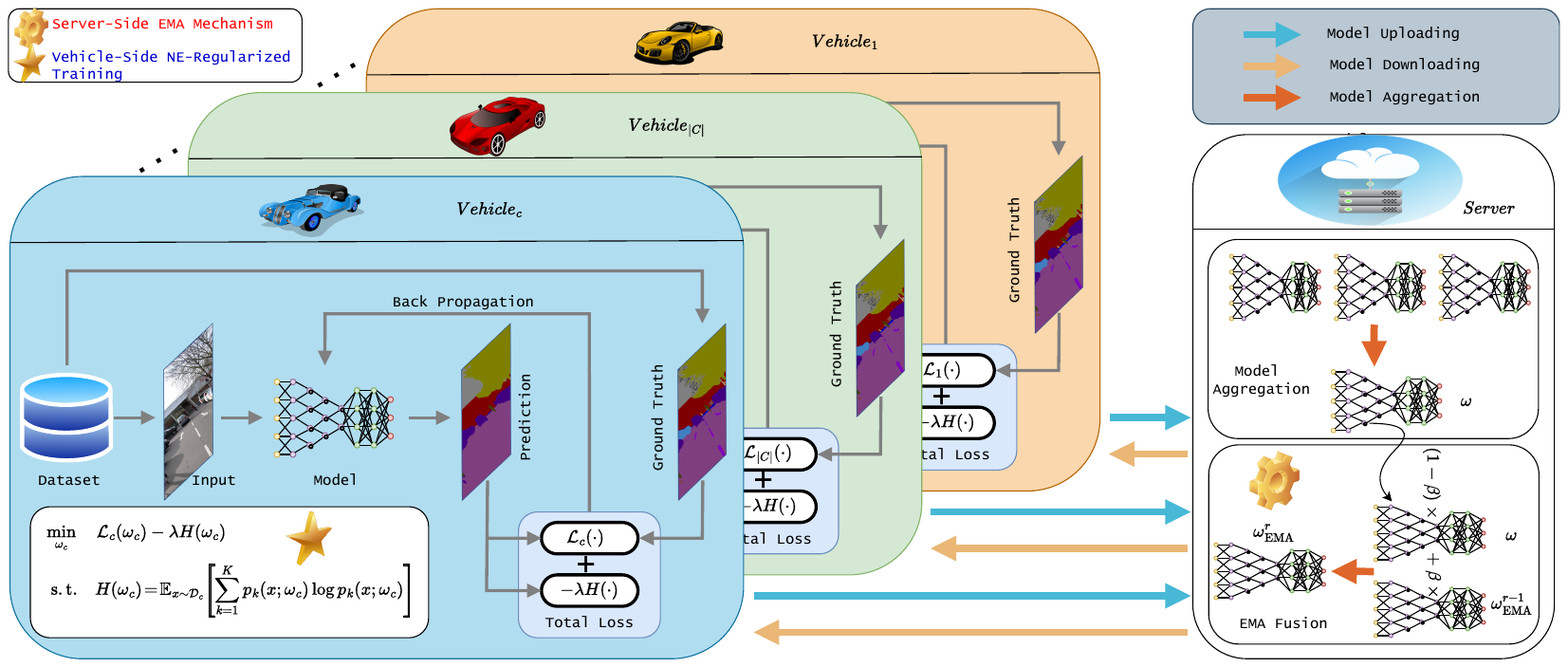}
\vspace{-6.5cm}
\caption{Overview of the proposed FedEMA method.}
\label{fig:FedEMA_framework}
\vspace{-0.7cm}
\end{figure*}

To mitigate such challenges, we propose \underline{F}ederated \underline{E}xponential \underline{M}oving \underline{A}verage (FedEMA), a novel framework that addresses these challenges through two synergistic strategies. First, an exponential moving average (EMA) approach with adjustable window size is proposed to fuse the currently aggregated model and the historical model. Specifically, in each FL round, an EMA model on the server is computed by weighted averaging current round's aggregated model and last round's EMA model in a recursive way. Afterwards, the current round's EMA model is distributed to all participated vehicles for next round's training. By doing so, local models of all involved vehicles preserve historical model's fitting capabilities while adapting to new patterns. As a result, it can effectively avoid short-term bias and fuse historical model's fitting abilities. Second, a negative entropy regularizer is proposed to be added to the training loss on each vehicle to penalize the local model complexity, which can mitigate the overfitting to temporal patterns introduced by the historical EMA model. These two integral innovations enable a dual-objective optimization to balance model generalization and adaptability. In addition, we offer theoretical convergence analysis for the proposed FedEMA to indicate how fast it converges and what factors influence the convergence. The proposed FedEMA is illustrated in \Cref{fig:FedEMA_framework}. Extensive evaluations on both Cityscapes dataset \cite{Cordts2016Cityscapes} and Camvid dataset \cite{brostow2008segmentation} demonstrate FedEMA’s superiority over existing state-of-the-art (SOTA) approaches. 

The main contributions of this work are highlighted as follows:
\begin{enumerate}
    \item We identify the catastrophic forgetting as a critical barrier to lifelong FedAD systems. To mitigate this issue, we propose FedEMA scheme to preserve historical model's fitting capabilities to overcome the catastrophic forgetting issue. In addition, to avoid the overfitting of temporal patterns introduced by historical EMA model, we propose to use negative entropy as a regularizer to penalize FedAD local model complexity.
    \item We additionally perform theoretical convergence analysis for the proposed FedEMA. It suggests that FedEMA's convergence rate $\mathcal{O}(1/\sqrt{R})$ matches standard federated optimization. It also indicates that the entropy regularizer improves the model generalization and accelerates the model convergence.
    \item Extensive experiments on both Cityscapes dataset and CamVid dataset demonstrate that the proposed FedEMA outperforms existing SOTA methods up to 7.12\% in mean Intersection over Union (mIoU).
\end{enumerate}

\section{Related Work}
\label{related_work}

\subsection{Federated Autonomous Driving (FedAD) System}
Current AD systems are generally categorized into two types: modular-based \cite{jiang2019multi} and learning-based \cite{10494721,10802211,10372140}. Modular-based methods, which are well-structured, suffer from error propagation due to potential inaccuracies in both their modeling and problem-solving phases. In contrast, learning-based end-to-end approaches \cite{9165167} represent a promising alternative by mitigating error propagation. These methods convert sensory inputs, such as LiDAR point clouds and camera imagery, directly into vehicle control actions, such as throttle, brake, and steering commands. Additionally, learning-based models can be integrated within modular pipelines, for instance, employing a learning-based model for semantic segmentation task within the semantic segmentation module \cite{10414408}. However, the primary challenge with learning-based approaches is their limited generalization capabilities, which can restrict performance to specific scenarios. 

Recently, Federated Learning (FL) is emerging as a solution to enhance the generalization of learning-based AD models \cite{dong2022federated}.  FL is a decentralized and distributed machine learning paradigm that emphasizes the data privacy preservation. It necessitates communication-efficient methods \cite{10529194,lin2023communication} to minimize communication overhead and enhance convergence rate. Initially, FedAvg \cite{https://doi.org/10.48550/arxiv.1602.05629} aggregates model parameters across vehicles using weighted averaging at a centralized server. However, researches have shown that data heterogeneity can adversely affect the convergence rate \cite{wang2021addressing,huang2021personalized}. To mitigate these issues, various strategies have been developed \cite{li2020federated,acar2021federated}. For instance, FedProx \cite{li2020federated} incorporates a proximal regularization term to local models, ensuring that the updated local parameters stay aligned with the global model, thus preventing gradient divergence. Similarly, FedDyn \cite{acar2021federated} introduces a dynamic regularizer for each device to synchronize global and local objectives. Although these FL methods improve the generalization of AD models to some extent, they typically encounter catastrophic forgetting issue due to their overwriting training approach. In this paper, we introduce the FedEMA to mitigate the catastrophic forgetting problem, thereby enhancing the generalization of the FedAD model in dynamic environments.

\subsection{Street Scene Semantic Understanding (S3U)}
Street Scene Semantic Understanding (S3U) \cite{kou2024adverse,kou2025label} is an interdisciplinary field bridging computer vision and AD, aimed at equipping AD vehicles with the ability to interpret and analyze the contents of street scenes. This analysis is typically achieved through processing various forms of sensory data, such as images and videos, which are vital for AD \cite{10342102}. S3U involves assigning a class label to every pixel in an image, a process essential for accurately depicting the layout of the street scene, including roads, pedestrians, sidewalks, buildings, and other elements, both static and dynamic. The field of S3U has evolved significantly with the integration of machine learning (ML), especially through the adoption of deep learning (DL) techniques. Initially, models based on Fully Convolutional Networks (FCNs) marked a substantial improvement in performance for these tasks\cite{zhou2022rethinking}. More recently, Transformer-based methods\cite{10341519} have been introduced for enhanced semantic segmentation. Additionally, the Bird's Eye View (BEV) technique\cite{9697426} has become increasingly popular for comprehensive road scene semantic understanding. In general, S3U is complex but crucial for AD and the proposed FedEMA can effectively balance S3U model's generalization and adaptability in real driving dynamic scenarios.

\section{Methodology}
\label{methodology}
We firstly formulate conventional FedAD system in \Cref{FedAD_formulation}. We then elaborate the proposed FedEMA method and explain why FedEMA can overcome the catastrophic problem of FedAD in \Cref{method_FedEMA_framework}. Finally, we conduct convergence analysis for the proposed FedEMA in \Cref{FedEMA_convergence}.

\subsection{FedAD Formulation}
\label{FedAD_formulation}
A FedAD system utilizes FL to collaboratively train AD model across multiple vehicles without sharing raw data. This decentralized approach allows vehicles to learn from collective experiences, leveraging diverse, real-world driving scenarios to refine AD model's generalization. Therefore, FedAD system is a promising solution to enhance the generalization of S3U model owing to its complexity and environmental variability.

\begin{table}[t]
    \centering
    \renewcommand{\arraystretch}{1.0}
    \setlength{\tabcolsep}{15.0pt}
    \caption{Key Notations of FedEMA Formulation}
    \begin{tabularx}{\linewidth}{ll}
    \hline
        \textbf{Symbols} & \textbf{Definitions} \\ \hline
        $c$ & Vehicle ID \\ 
        $\mathcal{C}$ & Vehicle set \\ 
        $\tau$ & Local update interval between two adjacent rounds \\ 
        $r$ & Round number of model aggregation \\ 
        $\mathcal{D}_c$ & Local dataset on Vehicle $c$ \\  
        $\omega_c$ & Local model parameters of Vehicle $c$ \\  
        $\omega$ & Global aggregated model parameters on the server \\  
        $p_c$ & Aggregation weight for Vehicle $c$ ($p_c = |\mathcal{D}_c|/|\mathcal{D}|$) \\  
        $R$ & Total aggregation rounds \\
        $\eta$ & Learning rate \\
        $d$ & Model dimension \\
        $\beta$ & EMA decay factor (0 < $\beta$ < 1) \\
        $\lambda$ & Negative entropy regularization coefficient \\
        $H(\cdot)$ & Negative entropy function \\
        $\omega_{\text{EMA}}^r$ & EMA model at round $r$ \vspace{0.04cm} \\
        \hline
    \end{tabularx}
\label{tab:FedEMA_Notations}
\vspace{-0.5cm}
\end{table}

The key notations in FedAD are listed in \Cref{tab:FedEMA_Notations}. We consider a FedAD system consisting of a central server and $|\mathcal{C}|$ vehicles. Vehicle $c$ has a local dataset $\mathcal{D}_c$ with size $|\mathcal{D}_c|$. The global dataset is defined as $\mathcal{D} \triangleq \cup_{c=1}^{|\mathcal{C}|} \mathcal{D}_c$ with total size $|\mathcal{D}| = \sum_{c=1}^{|\mathcal{C}|} |\mathcal{D}_c|$. The training of the conventional FedAD system iterates multiple rounds until convergence. For each round, it can be divided into following two steps:

\subsubsection{Vehicle Local Training}
In each local iteration, Vehicle $c$ trains its model $\omega_c$ using its local dataset $\mathcal{D}_c$. The local optimization objective is
\begin{equation}
\label{eq:vehicle_loss}
\min_{\omega_c} \mathcal{L}_c(\omega_c) = \frac{1}{|\mathcal{D}_c|}
\sum_{j=1}^{|\mathcal{D}_c|}\mathcal{E}(\omega_c, \mathcal{D}_c^{(j)}),
\end{equation}
where $\mathcal{E}(\omega_c, \mathcal{D}_c^{(j)})$ is the loss for training sample $j$. When Vehicle $c$ trains its local model for $\tau$ iterations, it sends the local optimized model to the central server for global model aggregation.

\subsubsection{Server Aggregation}
When the central server receives models from all participated vehicles, 
it performs global model aggregation, \ie,
\begin{align}
    \omega = \sum_{c=1}^{|\mathcal{C}|} p_c\omega_c, ~~~
    \mathcal{L}(\omega) = \sum_{c=1}^{|\mathcal{C}|} p_c\mathcal{L}_c(\omega),
\end{align}
where aggregation weight typically follows $p_c = |\mathcal{D}_c|/|\mathcal{D}|$. When the model aggregation is done, the server broadcasts the aggregated global model to all involved vehicles for next round's training.

With iterating $R$ aforementioned rounds, the FedAD system aims to find optimal global model $\omega$ that minimizes
\begin{align}
\label{eq:global_objective}
\min_{\omega \in \mathbb{R}^d} \mathcal{L}(\omega) & = \sum_{c=1}^{|\mathcal{C}|} p_c\mathcal{L}_c(\omega), \quad 
\mathrm{s.t.} \quad \omega = \sum_{c=1}^{|\mathcal{C}|} p_c\omega_c.
\end{align}

\subsection{The Proposed FedEMA} \label{method_FedEMA_framework}
Recall that the conventional FedAD systems face the challenge of catastrophic forgetting owing to its overwritting training method. To overcome this issue, we propose the FedEMA method. The proposed FedEMA extends standard FedAD system with following two integral components. 
\begin{itemize}
    \item \textbf{Server-Side Exponential Moving Averaging Mechanism:} The central server maintains an EMA model that blends current round's aggregated model and last round's EMA model. Specifically, taking the $r$-th round as example, apart from the global model aggregation in the conventional FedAD, we propose to further fuse the aggregated global model $\omega$ and the last round's EMA model to compute current round's EMA model. This summation is formulated as
\begin{align}
    \omega_{\text{EMA}}^r = \beta\omega_{\text{EMA}}^{r-1} + (1-\beta)\omega, \label{eq:ema_update}
\end{align}
where $\beta = \frac{2}{N+1}$ and $N$ means the considered window size in the proposed FedEMA. After $\omega_{\text{EMA}}^r$ is computed, it is distributed to all participated vehicles for next round's training. Through this EMA model fusion strategy, the trained model contains the fitting capabilities of historical models, thereby overcoming the catastrophic forgetting problem in conventional FedAD system.

\item \textbf{Vehicle-Side Negative Entropy-Regularized Training:}
For each vehicle, apart from the original training loss $\mathcal{L}_c(\cdot)$, we propose to use negative entropy $H(\cdot)$ as a regularizer to penalize the model complexity to avoid temporal overfitting owing to the consideration of historical EMA model. Taking Vehicle $c$ as an example, it optimizes a modified local objective with negative entropy regularization, \ie,
\begin{equation}
\label{eq:fedema_vehicle}
\min_{\omega_c} \mathcal{L}_c(\omega_c) - \lambda H(\omega_c),
\end{equation}
where $H(\omega_c)\!=\! \mathbb{E}_{x\sim\mathcal{D}_c}\!\left[\sum_{k=1}^K p_k(x;\omega_c)\log p_k(x;\omega_c)\right]$ measures the negative entropy of the output $p_k(x;\omega_c)$, which is the softmax probability for class $k$ at pixel $x$. By maximizing the negative entropy regularizer, it can enhance model generalization by preventing the overconfident predictions.
\end{itemize}

With aforementioned two extensions, the optimization problem of the proposed FedEMA is to find the optimal EMA model, \ie,
\begin{align}
\begin{aligned}
\min_{\omega_{\text{EMA}}^r} & \quad \sum_{c=1}^{|\mathcal{C}|} p_c\left[\mathcal{L}_c(\omega_{\text{EMA}}^r) - \lambda H(\omega_{\text{EMA}}^r)\right] \\
\mathrm{s.t.} & \quad \omega_{\text{EMA}}^r = \beta\omega_{\text{EMA}}^{r-1} + (1-\beta)\sum_{c=1}^{|\mathcal{C}|} p_c\omega_c, \\
& \quad \omega_c = \arg\min_{\omega} \left[\mathcal{L}_c(\omega) + \lambda H(\omega)\right].
\end{aligned}
\label{eq:FedEMA_optim}
\end{align}
Notably, when optimization problem \Cref{eq:FedEMA_optim} is solved, $\omega_{\text{EMA}}^r$ instead of aggregated model $\omega$ servers the target global model in FedAD system. 

The training of the proposed FedEMA starts with all involved vehicles' models' initialization by $\omega_{\text{EMA}}^0$. After initialization, the training iterates $R$ rounds. For $r$-th round, where $r \in \{1, 2, \cdots, R\}$, it follows below steps:
\begin{itemize}
    \item \textbf{Vehicle Update}: Each vehicle performs $\tau$ local negative entropy-regularized updates, \ie,
    \begin{equation}
    \omega_c = \omega_c - \eta\left[\nabla\mathcal{L}_c(\omega_c) - \lambda\nabla H(\omega_c)\right].
    \end{equation}

    \item \textbf{Federated Aggregation}: The server receives models from all participated vehicles and aggregates these models to generate aggregation model $\omega$.
    
    \item \textbf{EMA Aggregation}: The server fuses the aggregated model $\omega$ and the $(r-1)$-th round's EMA model $\omega_{\text{EMA}}^{r-1}$ to generate $\omega_{\text{EMA}}^{r}$ using \Cref{eq:ema_update}.
    
    \item \textbf{Model Distribution}: The server distributes $\omega_{\text{EMA}}^r$ to all participated vehicles.
\end{itemize}

In summary, the training process of the proposed FedEMA is outlined in \Cref{alg:fedema}.

\setlength{\textfloatsep}{0pt}
\begin{algorithm}[t]
\caption{FedEMA}
\label{alg:fedema}
\KwIn{Vehicle set $\mathcal{C}$, initial model $\omega^0$, EMA factor $\beta$, regularizer coefficient $\lambda$, local update interval $\tau$, training rounds $R$}
\KwOut{Global EMA model $\omega_{\text{EMA}}^R$}

Initialize $\omega_{\text{EMA}}^0 \leftarrow \omega^0$\;
\For{round $r = 1, 2, \ldots, R$}{
    \tcp{Server executes}
    Send $\omega_{\text{EMA}}^{r-1}$ to all vehicles $c \in \mathcal{C}$\;
    
    \tcp{Vehicle update in parallel}
    \ForEach{vehicle $c \in \mathcal{C}$}{
        $\omega_c \leftarrow \omega_{\text{EMA}}^{r-1}$\;
        \For{local step $t = 0, \ldots, \tau-1$}{
            Update local model: $\omega_c \leftarrow \omega_c - \eta \nabla\left[\mathcal{L}_c(\omega_c) - \lambda H(\omega_c)\right]$\;
        }
        Send $\omega_c$ to server\;
    }
    
    \tcp{Federated and EMA aggregation}
    Compute aggregated model: 
    $\omega \leftarrow \sum_{c\in\mathcal{C}} \frac{|\mathcal{D}_c|}{|\mathcal{D}|} \omega_c$\;
    Update EMA model: 
    $\omega_{\text{EMA}}^r \leftarrow \beta\omega_{\text{EMA}}^{r-1} + (1-\beta)\omega$\;
}
\end{algorithm}

\subsection{FedEMA Convergence Analysis}
\label{FedEMA_convergence}
With respect to S3U task, the local loss $\mathcal{L}_c(\omega)$ is the cross-entropy (CE) loss. For vehicle $c$ with dataset $\mathcal{D}_c$, the CE loss is formulated as 
\begin{align}
\mathcal{L}_c(\omega) = -\frac{1}{|\mathcal{D}_c|}\sum_{(x,y)\in\mathcal{D}_c} \sum_{k=1}^K y_k \log p_k(x;\omega_c),
\end{align}
where $y_k$ is the one-hot label for class $k$ at pixel $x$, and $p_k(x;\omega_c)$ is the softmax probability for class $k$ at pixel $x$.

To conduct FedEMA convergence analysis, some assumptions are made as follows:
\begin{itemize}
    \item \textbf{L-smoothness}: For all $\omega, \omega'$, and $c$, $\|\nabla\mathcal{L}_c(\omega) - \nabla\mathcal{L}_c(\omega')\| \leq L\|\omega - \omega'\|$.
    \item \textbf{Bounded gradients}: $\mathbb{E}\|\nabla\mathcal{L}_c(\omega)\|^2\!\leq\!G^2$ for all $c, \omega$.
    \item \textbf{Bounded heterogeneity}: $\mathbb{E}_c\|\nabla\mathcal{L}_c(\omega) - \nabla\mathcal{L}(\omega)\|^2 \leq \sigma^2$, where $\mathcal{L}(\omega) = \sum_c p_c \mathcal{L}_c(\omega)$.
    \item \textbf{Negative Entropy regularity}: The negative entropy $H(\omega)$ satisfies $\|\nabla H(\omega)\| \leq \gamma\mathbb{E}_x\|\nabla \log p_k(x;\omega)\|$.
\end{itemize}

Based on these assumptions, we can conclude following \textbf{Theorem 1} about the convergence of the proposed FedEMA.

\begin{theorem}
Under non-convex loss setting, after $R$ rounds' training with learning rate $\eta = \frac{1}{\sqrt{R}}$, we have
\begin{align}
\!\frac{1}{R}\!\sum_{r=1}^R\!\mathbb{E}\|\nabla\mathcal{L}(\omega_{\text{EMA}}^r)\|^2\!\leq\! \mathcal{O}\!\left(\!\frac{\mathcal{L}(\omega^0)\!-\!\mathcal{L}^*}{\sqrt{R}}\!+\!\frac{G^2\!+\!\lambda^2\gamma^2\!+\!\sigma^2}{\sqrt{R}}\right),
\end{align}
\label{theorem_1}
where $\mathcal{L}^*$ is the theoretical optimal loss and can be treated as a constant, and $\omega^0$ is the initial model of the training (\ie, $\omega^0 = \omega^0_{\text{EMA}}$).
\end{theorem}

From \textbf{Theorem 1}, we can conclude following insights: (I) The convergence rate $\mathcal{O}(1/\sqrt{R})$ matches standard federated optimization bounds. (II) The entropy regularization term $\lambda\gamma$ directly reduces client drift, improving generalization. (III) Larger $\lambda$ suppresses heterogeneity $\sigma^2$, which leads to faster convergence.

This theorem can be proved by following steps:

\textbf{Step 1: Reformulate EMA as Momentum.}  
The server-side EMA update can be rewritten as
\begin{align}
\omega_{\text{EMA}}^r - \omega_{\text{EMA}}^{r-1} = (1-\beta)\left(\sum_c p_c\omega_c - \omega_{\text{EMA}}^{r-1}\right),
\end{align}
which mirrors Nesterov momentum with factor $\beta$.

\textbf{Step 2: Bound Client Drift.}  
Let $\omega_c^{r,t}$ denote the $t$-th (where $t \in \{1, 2, \cdots, \tau\}$) local step of vehicle $c$ in round $r$. The local update with entropy regularization satisfies
\begin{align}
\!\!\|\omega_c^{r,\tau}\!-\!\omega_{\text{EMA}}^{r-1}\| \!\leq\! \eta\tau\left(G\!+\!\lambda\gamma\right)\!+\!\mathcal{O}(\eta^2\tau^2 L(G^2\!+\!\lambda^2\gamma^2)).
\end{align}

\textbf{Step 3: Descent Lemma for CE Loss.}  
Using L-smoothness and telescoping over rounds,
\begin{align}
\!\mathcal{L}(\omega_{\text{EMA}}^r)\!\leq\!&\mathcal{L}(\omega_{\text{EMA}}^{r-1})\!-\!\eta(1\!-\!\beta)\!\left\langle\! \nabla\mathcal{L}(\omega_{\text{EMA}}^{r-1}),\!\sum_c p_c\nabla\mathcal{L}_c(\omega_c^{r,\tau})\!\right\rangle \notag \\ &+ \frac{\eta^2 L(1-\beta)^2}{2}\mathbb{E}\left\|\sum_c p_c\nabla\mathcal{L}_c(\omega_c^{r,\tau})\right\|^2.
\end{align}

\textbf{Step 4: Convergence Rate.}  
Under non-convex losses, after $R$ rounds' training with learning rate $\eta = \frac{1}{\sqrt{R}}$,
\begin{align}
\!\frac{1}{R}\!\sum_{r=1}^R\!\mathbb{E}\|\nabla\mathcal{L}(\omega_{\text{EMA}}^r)\|^2\!\leq\! \mathcal{O}\!\left(\!\frac{\mathcal{L}(\omega^0)\!-\!\mathcal{L}^*}{\sqrt{R}}\!+\!\frac{G^2\!+\!\lambda^2\gamma^2\!+\!\sigma^2}{\sqrt{R}}\right),
\end{align}
where $\mathcal{L}^*$ is the theoretical optimal loss and can be treated as a constant, and $\omega^0$ is the initial model of the training (\ie, $\omega^0 = \omega^0_{\text{EMA}}$).

\section{Experiments}
\label{experiments}
This section details experiments on the S3U task. We aim to measure the enhancement of FedAD model generalization against mitigating the catastrophic forgetting problem, employing widely recognized and accepted metrics.

\subsection{Datasets, Evaluation Metrics, and Implementation}
\subsubsection{Datasets}
The \textbf{Cityscapes} dataset \cite{Cordts2016Cityscapes} comprises 2,975 training images and 500 validation images, each annotated with masks. This dataset encompasses 19 semantic classes, including vehicles and pedestrians. For our experiments, the training data is distributed among FedAD vehicles. The \textbf{CamVid} dataset \cite{brostow2008segmentation} consists of 701 samples across 11 semantic classes. We randomly selected 600 samples to distribute among FedAD vehicles for training, reserving the remaining 101 samples as a test dataset. 

\begin{table}[tp]
\vspace{-0.3cm}
    \centering
    \renewcommand{\arraystretch}{1.0}
    \setlength{\tabcolsep}{15.0pt}
    \caption{Hardware/Software configurations}
    \begin{tabularx}{\linewidth}{ll}
    \hline
        \textbf{Items} & \textbf{Configurations} \\ \hline
        CPU  & AMD Ryzen 9 3900X 12-Core \\ 
        GPU  & NVIDIA GeForce 3090 $\times$ 2\\ 
        RAM  & DDR4 32G \\ 
        DL Framework  & PyTorch @ 1.13.0+cu116 \\ 
        GPU Driver  & 470.161.03 \\ 
        CUDA  & 11.4 \\ 
        cuDNN  & 8302 \\ \hline
    \end{tabularx}
\label{Tab:configs}
\end{table}

\begin{table}[tp]
\vspace{-0.3cm}
    \centering
    \renewcommand{\arraystretch}{1.0}
    \setlength{\tabcolsep}{15.0pt}
    \caption{Training configurations}
    \begin{tabularx}{\linewidth}{ll}
    \hline
        \textbf{Items} & \textbf{Configurations} \\ \hline
        Loss  & nn.CrossEntropyLoss \\ 
        Optimizer  & nn.Adam \\ 
        Adam Betas  & (0.9, 0.999) \\ 
        Weight Decay  & 1e-4 \\ 
        Batch Size  & 8 \\ 
        Learning Rate  & 3e-4 \\ 
        \multirow{1}{*}{DNN Models}  & DeepLabv3+ \cite{chen2018encoderdecoder}, TopFormer \cite{zhang2022topformer} \\
        \multirow{2}{*}{FL Algorithms} &FedProx \cite{li2020federated}, FedDyn \cite{acar2021federated} \\
        ~ &FedIR \cite{hsu2020federated}, MOON \cite{li2021model}\\
        \hline
    \end{tabularx}
\label{Tab:train}
\end{table}

\subsubsection{Evaluation Metrics}
We evaluate the proposed FedEMA on S3U task using four metrics: \textbf{mIoU}, \textbf{mPrecision (abbreviated as mPre)}, \textbf{mRecall (abbreviated as mRec)}, and \textbf{mF1}. These metrics are formally defined as follows:
\vspace{-0.2cm}
\begin{align}
    &\!mIoU\!=\!\frac{1}{\mathcal{K}}\!\sum_{k=1}^{\mathcal{K}}\!IoU_k\!=\!\frac{1}{\mathcal{K}\!*\!\mathcal{N}}\!\sum_{k=1}^{\mathcal{K}}\! \sum_{n=1}^{\mathcal{N}}\!\frac{TP_{n, k}}{FP_{n, k}\!+\!TP_{n, k}\!+\!FN_{n, k}},
    \nonumber
    \\
    \vspace{-0.2cm}
    &mPre = \frac{1}{\mathcal{K}}\sum_{k=1}^{\mathcal{K}}Pre_k = \frac{1}{\mathcal{K} * \mathcal{N}}\sum_{k=1}^{\mathcal{K}} \sum_{n=1}^{\mathcal{N}} \frac{TP_{n, k}}{FP_{n, k} + TP_{n, k}},
    \nonumber
    \\
    \vspace{-0.2cm}
    &mRec = \frac{1}{\mathcal{K}}\sum_{k=1}^{\mathcal{K}}Rec_k = \frac{1}{\mathcal{K} * \mathcal{N}}\sum_{k=1}^{\mathcal{K}} \sum_{n=1}^{\mathcal{N}} \frac{TP_{n, k}}{TP_{n, k} + FN_{n, k}},
    \nonumber
    \\
    \vspace{-0.2cm}
    &mF1 = \frac{1}{\mathcal{K}}\sum_{k=1}^{\mathcal{K}}F1_k = \frac{1}{\mathcal{K}}\sum_{k=1}^{\mathcal{K}} \frac{2 * Pre_k * Rec_k}{Pre_k + Rec_k},
    \label{Eq:mF1}
\end{align}
where $TP$, $FP$, $TN$, and $FN$ represent True Positive, False Positive, True Negative, and False Negative, respectively. The term $\mathcal{K}$ indicates the number of semantic classes in the test dataset, set at 19 for the Cityscapes dataset and 11 for the CamVid dataset. Additionally, $\mathcal{N}$ denotes the size of the test dataset, which is 500 for Cityscapes and 101 for CamVid.

\subsubsection{Implementation Details}
The primary hardware and software configurations used in our study are detailed in \Cref{Tab:configs}, while the training specifics are provided in \Cref{Tab:train}. Our experiments include a comparative analysis between the proposed FedEMA and several other FL algorithms listed in \Cref{Tab:train}. Some algorithms, such as FedDyn and FedProx, are evaluated with their respective hyperparameters, which are noted in brackets, \eg, FedDyn(0.01).

\begin{table*}[tp]
\centering
\setlength{\tabcolsep}{6.2pt}
\caption{Metrics on both Cityscapes and CamVid dataset driven by \textbf{DeepLabv3+} model}
\begin{tabularx}{\linewidth}{cccccccccc}
\hline
\multirow{2}{*}{FL Algorithms}                                   & \multicolumn{4}{c}{Cityscapes Dataset (19 Semantic Classes) (\%)}                                                                                         & & \multicolumn{4}{c}{CamVid Dataset (11 Semantic Classes) (\%)}              \\ \cline{2-5}  \cline{7-10}
                                                                 & mIoU                               & mF1                                & mPrecision                         & mRecall                          &  & mIoU           & mF1            & mPrecision     & mRecall        \\ \hline
FedDyn (0.005) \cite{acar2021federated} &41.58$\pm$2.83 &36.36$\pm$2.28 &47.35$\pm$1.46 &42.21$\pm$2.55 &&72.85$\pm$0.35 &80.28$\pm$0.26 &81.40$\pm$0.11 &79.37$\pm$0.37 \\
FedDyn (0.01) \cite{acar2021federated}  &35.49$\pm$0.33 &31.49$\pm$0.31 &34.91$\pm$0.16 &36.84$\pm$0.36 &&69.31$\pm$1.35 &77.10$\pm$1.51 &81.32$\pm$0.41 &75.71$\pm$1.42 \\
FedIR \cite{hsu2020federated}           &31.28$\pm$0.21 &27.96$\pm$0.24 &32.46$\pm$0.03 &31.09$\pm$0.28 &&61.37$\pm$0.87 &68.19$\pm$0.76 &76.73$\pm$0.12 &64.94$\pm$0.87 \\
FedProx (0.005) \cite{li2020federated}  &\underline{56.59$\pm$0.24} &\underline{48.41$\pm$0.23} &\underline{58.41$\pm$0.15} &\underline{56.78$\pm$0.48} &&\underline{74.60$\pm$0.10} &\textbf{82.18$\pm$0.07} &\underline{82.66$\pm$0.11} &\textbf{81.78$\pm$0.11} \\
FedProx (0.01) \cite{li2020federated}   &47.47$\pm$0.18 &41.41$\pm$0.21 &47.47$\pm$0.11 &48.07$\pm$0.29 &&74.31$\pm$0.19 &81.29$\pm$0.13 &82.15$\pm$0.15 &80.58$\pm$0.21 \\
MOON \cite{li2021model}                 &53.79$\pm$0.18 &45.94$\pm$0.20 &55.53$\pm$0.13 &53.33$\pm$0.30 &&72.08$\pm$0.68 &79.34$\pm$0.99 &\textbf{84.16$\pm$0.55} &77.71$\pm$0.65 \\
\textbf{FedEMA (Ours)}                  &\textbf{60.62$\pm$1.38} &\textbf{51.60$\pm$1.23} &\textbf{64.33$\pm$1.17} &\textbf{59.17$\pm$0.86} &&\textbf{74.73$\pm$0.23} &\underline{81.59$\pm$0.15} &82.27$\pm$0.11 &\underline{81.02$\pm$0.21} \vspace{0.02cm} \\ \hline
\end{tabularx}
\label{Tab:metrics_deeplabv3}
\vspace{-0.4cm}
\end{table*}

\begin{table*}[tp]
\centering
\setlength{\tabcolsep}{6.2pt}
\caption{Metrics on both Cityscapes and CamVid dataset driven by \textbf{TopFormer} model}
\begin{tabularx}{\linewidth}{cccccccccc}
\hline
\multirow{2}{*}{FL Algorithms}                                   & \multicolumn{4}{c}{Cityscapes Dataset (19 Semantic Classes) (\%)}                                                                                         & & \multicolumn{4}{c}{CamVid Dataset (11 Semantic Classes) (\%)}              \\ \cline{2-5}  \cline{7-10}
                                                                 & mIoU                               & mF1                                & mPrecision                         & mRecall                          &  & mIoU           & mF1            & mPrecision     & mRecall        \\ \hline
FedDyn (0.005) \cite{acar2021federated} &19.30$\pm$0.09 &22.18$\pm$0.06 &20.84$\pm$0.07 &23.76$\pm$0.06 &&36.48$\pm$0.51 &43.13$\pm$0.41 &43.24$\pm$0.25 &43.34$\pm$0.51 \\
FedDyn (0.01) \cite{acar2021federated}  &18.89$\pm$0.08 &21.90$\pm$0.06 &20.52$\pm$0.05 &23.56$\pm$0.07 &&37.48$\pm$0.41 &43.93$\pm$0.34 &43.19$\pm$0.33 &44.89$\pm$0.29 \\
FedIR \cite{hsu2020federated}           &21.30$\pm$0.04 &23.52$\pm$0.03 &22.58$\pm$0.02 &24.68$\pm$0.04 &&40.44$\pm$0.56 &46.10$\pm$0.42 &47.88$\pm$0.22 &45.48$\pm$0.49 \\
FedProx (0.005) \cite{li2020federated}  &\underline{25.73$\pm$0.10} &\underline{29.49$\pm$0.08} &\underline{28.92$\pm$0.09} &\underline{30.48$\pm$0.11} &&\textbf{42.10$\pm$0.31} &\textbf{48.58$\pm$0.24} &\underline{48.88$\pm$0.29} &\underline{48.62$\pm$0.29} \\
FedProx (0.01) \cite{li2020federated}   &25.06$\pm$0.21 &28.92$\pm$0.20 &28.74$\pm$0.12 &29.71$\pm$0.22 &&41.71$\pm$1.13 &47.57$\pm$1.57 &\textbf{51.34$\pm$3.73} &48.51$\pm$1.06 \\
MOON \cite{li2021model}                 &0.20 $\pm$0.54 &0.31 $\pm$0.73 &0.20 $\pm$0.50 &5.26 $\pm$1.75 &&1.03 $\pm$0.84 &1.70 $\pm$1.16 &2.92 $\pm$2.71 &7.23 $\pm$2.80 \\
\textbf{FedEMA (Ours)}                  &\textbf{25.93$\pm$0.21} &\textbf{29.84$\pm$0.20} &\textbf{29.56$\pm$0.10} &\textbf{30.63$\pm$0.23} &&\underline{41.79$\pm$0.36} &\underline{47.78$\pm$0.26} &47.18$\pm$0.27 &\textbf{48.87$\pm$0.23} \vspace{0.02cm} \\ \hline
\end{tabularx}
\label{Tab:metrics_topformer}
\vspace{-0.5cm}
\end{table*}

\begin{figure*}[tp]
\centering
\subfloat[\footnotesize mIoU]{\includegraphics[width=0.24\linewidth]{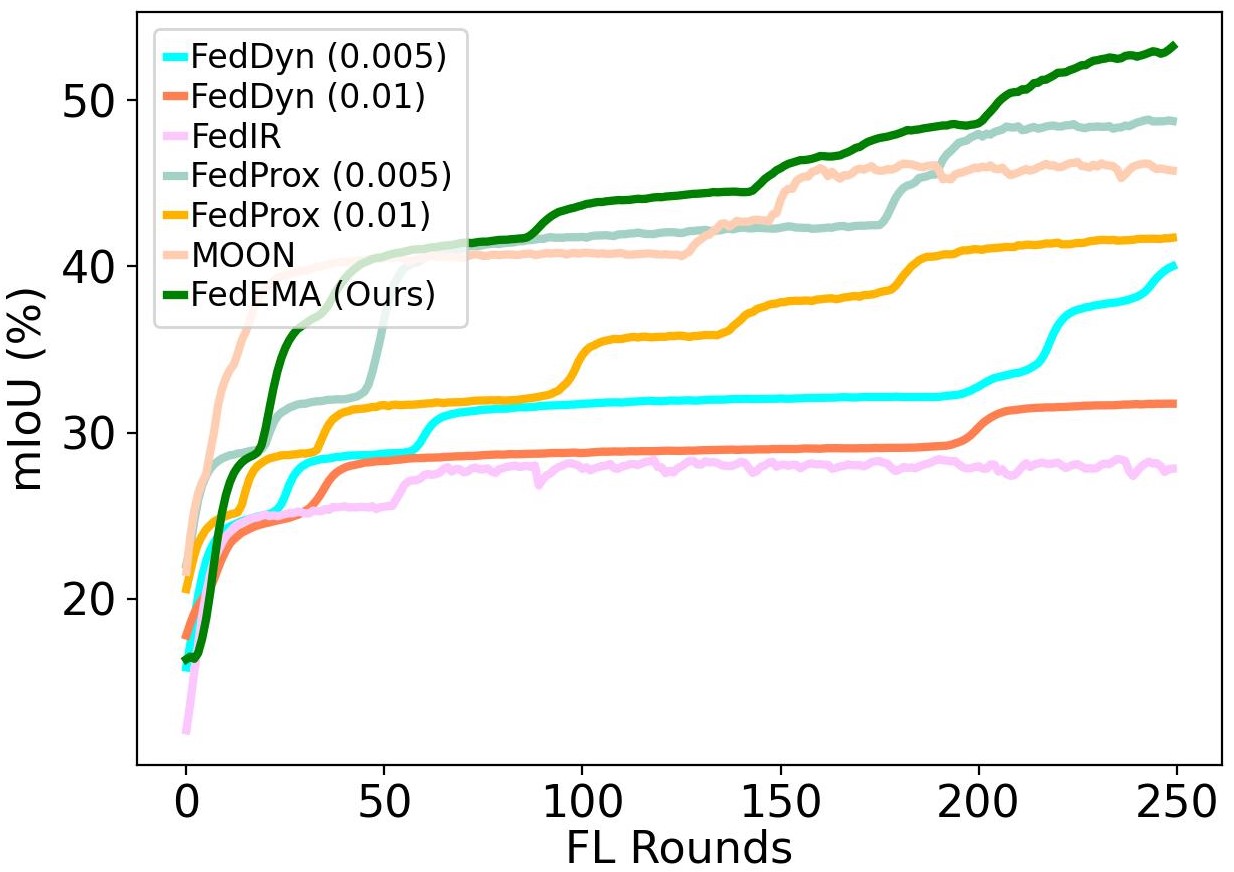}
\label{Fig.Metrics_mIoU}
}
\subfloat[\footnotesize mPrecision]{\includegraphics[width=0.24\linewidth]{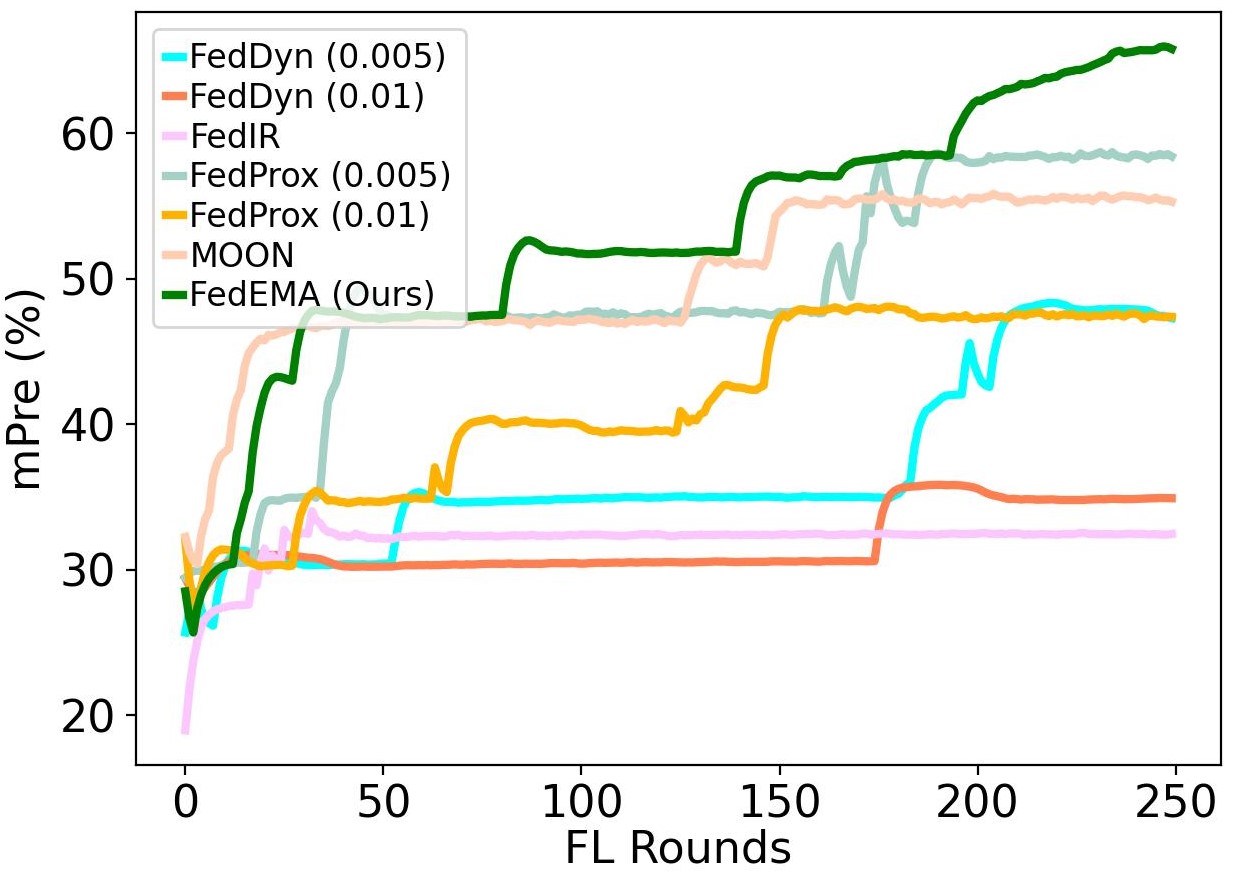}
\label{Fig.Metrics_mPre}
}
\subfloat[\footnotesize mRecall]{\includegraphics[width=0.24\linewidth]{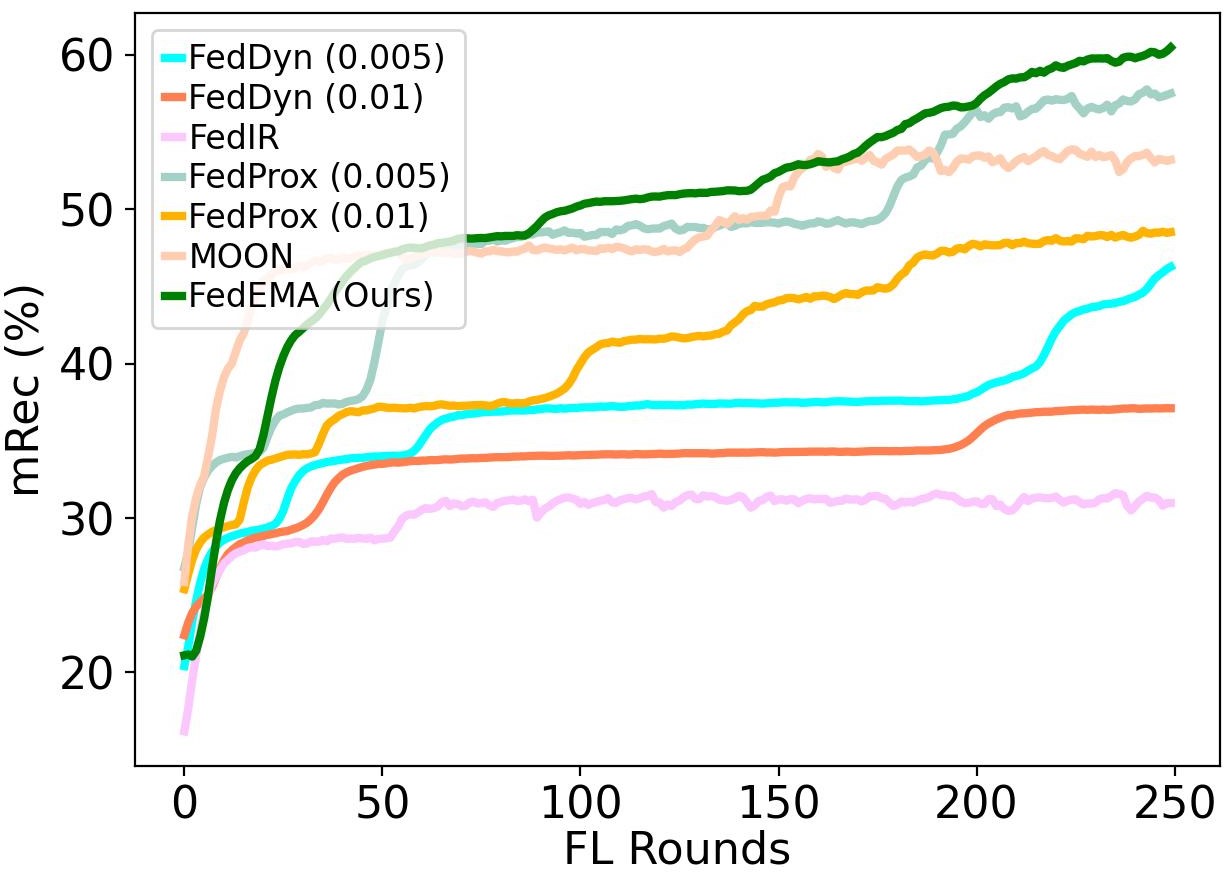}
\label{Fig.Metrics_mRec}
}
\subfloat[\footnotesize mF1]{\includegraphics[width=0.24\linewidth]{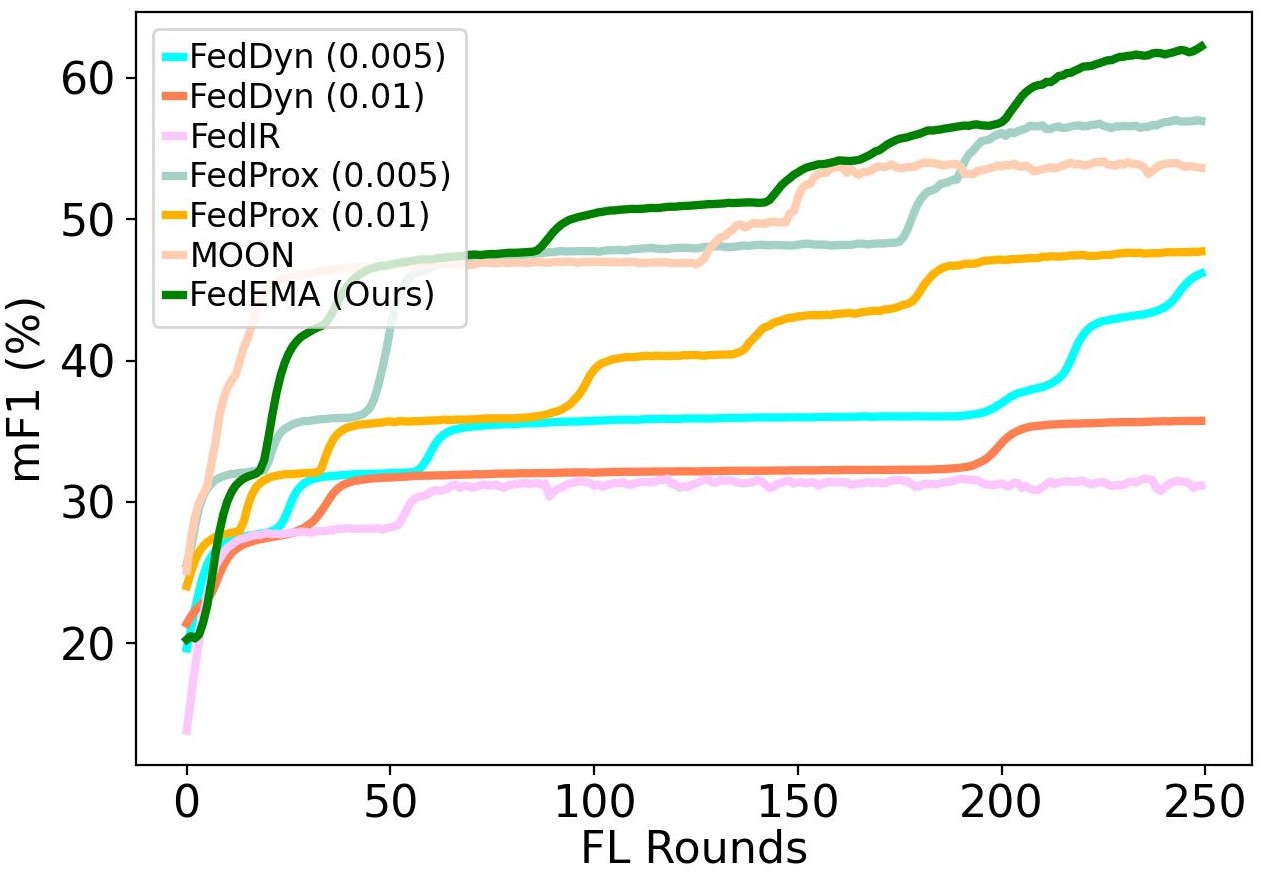}
\label{Fig.Metrics_mF1}
}
\vspace{-0.15cm}
\caption{Metric comparison of DeepLabv3+ on Cityscapes dataset. Results show FedEMA outperform other FL methods across all metrics.}
\label{Fig.Metrics}
\vspace{-0.6cm}
\end{figure*}

\subsection{Main Results and Empirical Analysis}

\subsubsection{Quantitative Performance Comparison}
\begin{table*}[tp]
\centering
\renewcommand{\arraystretch}{0.24}
\addtolength{\tabcolsep}{-0.45pt}
\caption{Prediction performance comparison of semantic understanding driven by varieties of FL algorithms}
\begin{tabularx}{\linewidth}{|l|lllll|}
\hline
\verticaltext[23pt]{Raw RGBs} &\includegraphics[width=0.187\linewidth, height=0.10\linewidth]{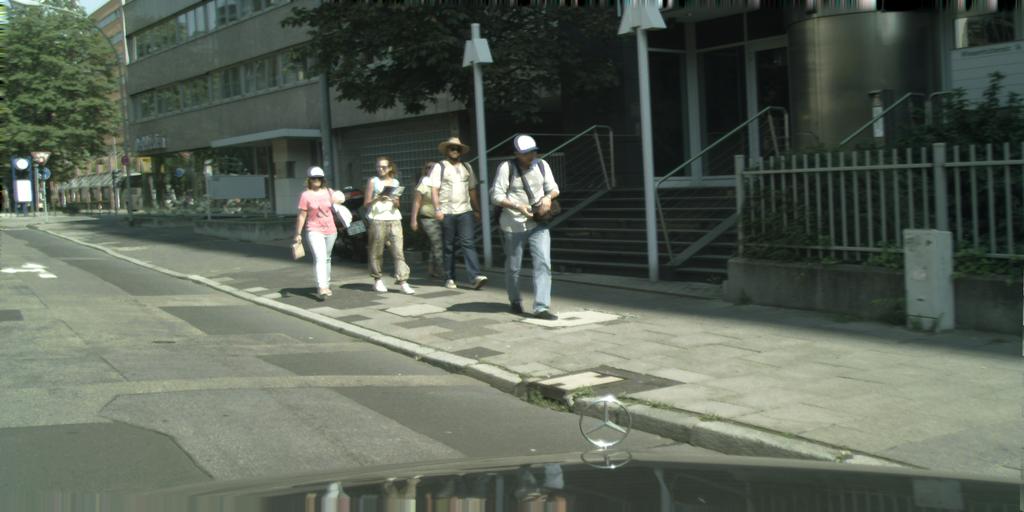} &\hspace{-0.47cm}
\includegraphics[width=0.187\linewidth, height=0.10\linewidth]{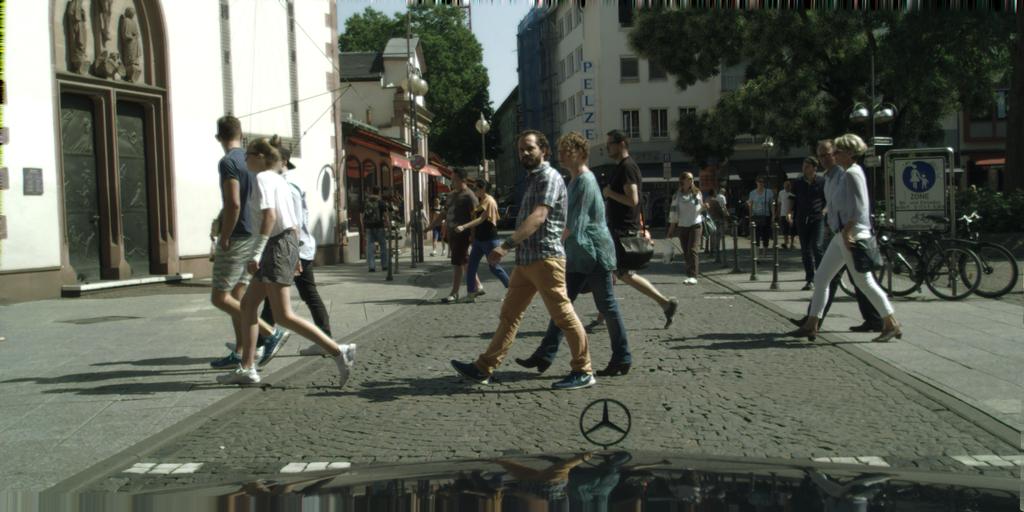} &\hspace{-0.47cm}
\includegraphics[width=0.187\linewidth, height=0.10\linewidth]{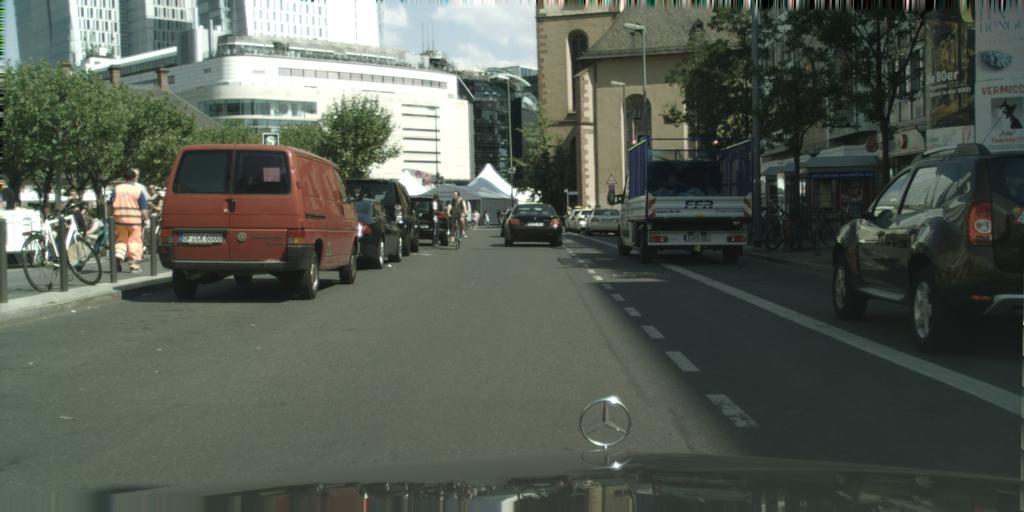} &\hspace{-0.47cm}
\includegraphics[width=0.187\linewidth, height=0.10\linewidth]{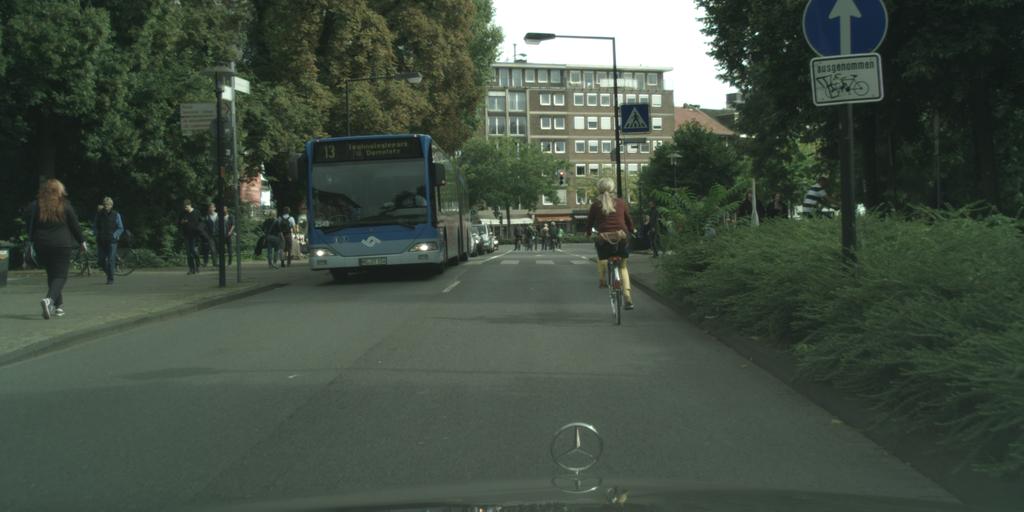} &\hspace{-0.47cm}
\includegraphics[width=0.187\linewidth, height=0.10\linewidth]{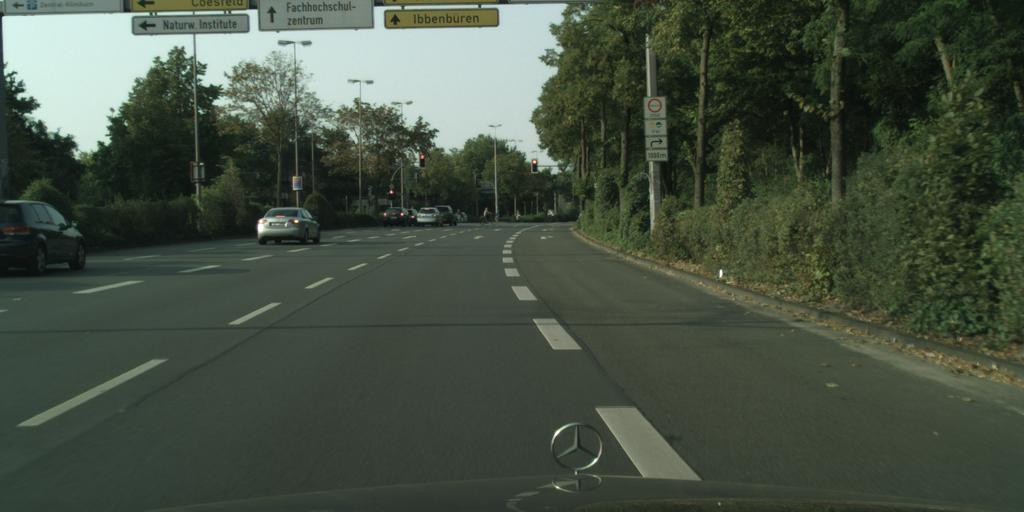}\\
\hline

\verticaltext[22pt]{Ground Truth} &
\includegraphics[width=0.187\linewidth, height=0.10\linewidth]{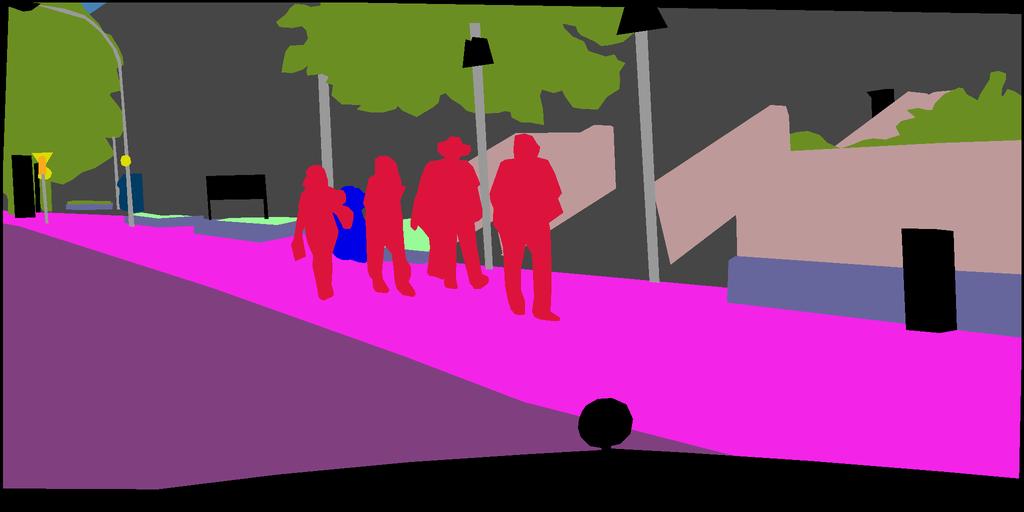} &\hspace{-0.47cm}
\includegraphics[width=0.187\linewidth, height=0.10\linewidth]{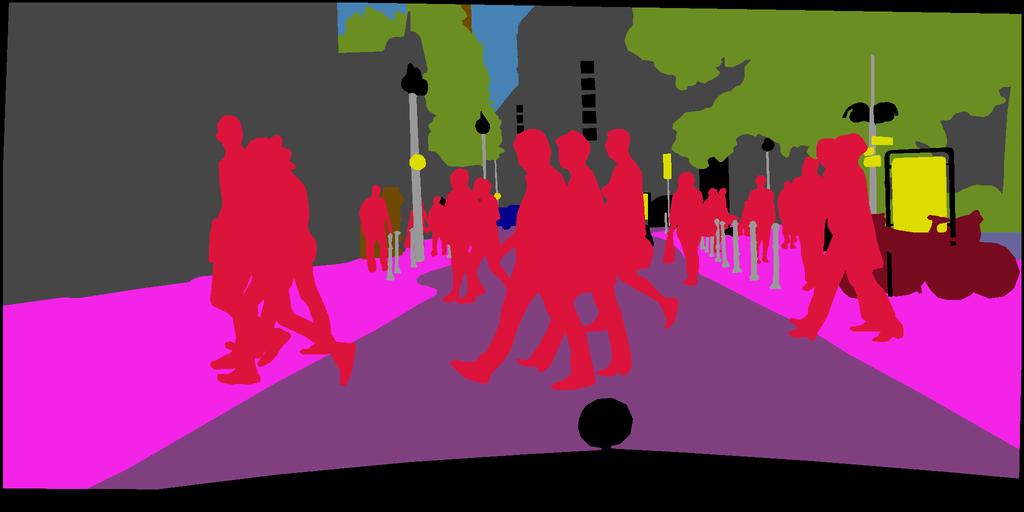} &\hspace{-0.47cm}
\includegraphics[width=0.187\linewidth, height=0.10\linewidth]{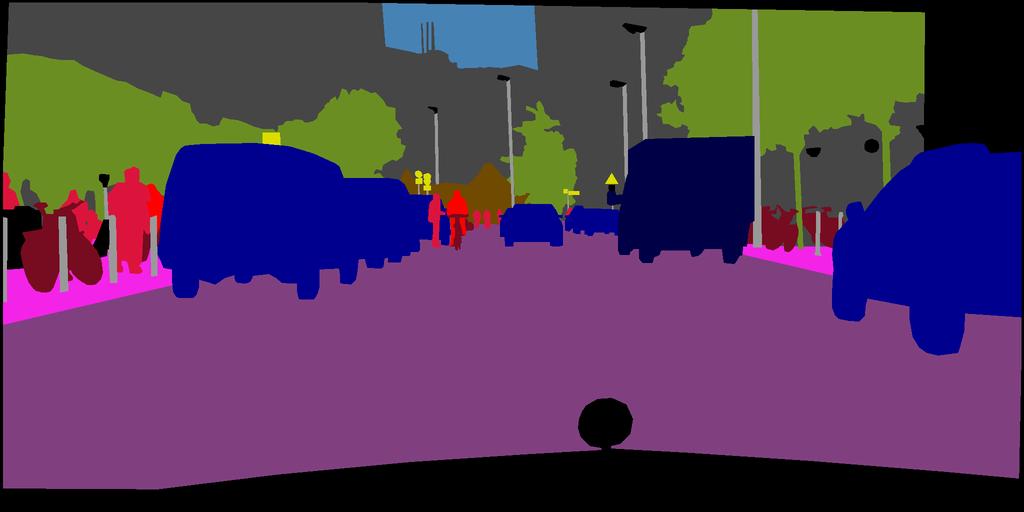} &\hspace{-0.47cm}
\includegraphics[width=0.187\linewidth, height=0.10\linewidth]{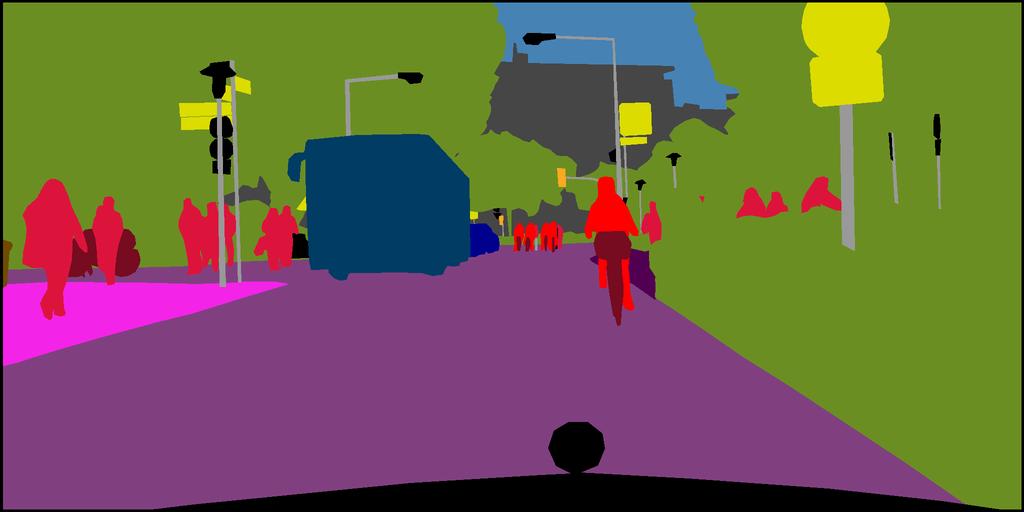  } &\hspace{-0.47cm}
\includegraphics[width=0.187\linewidth, height=0.10\linewidth]{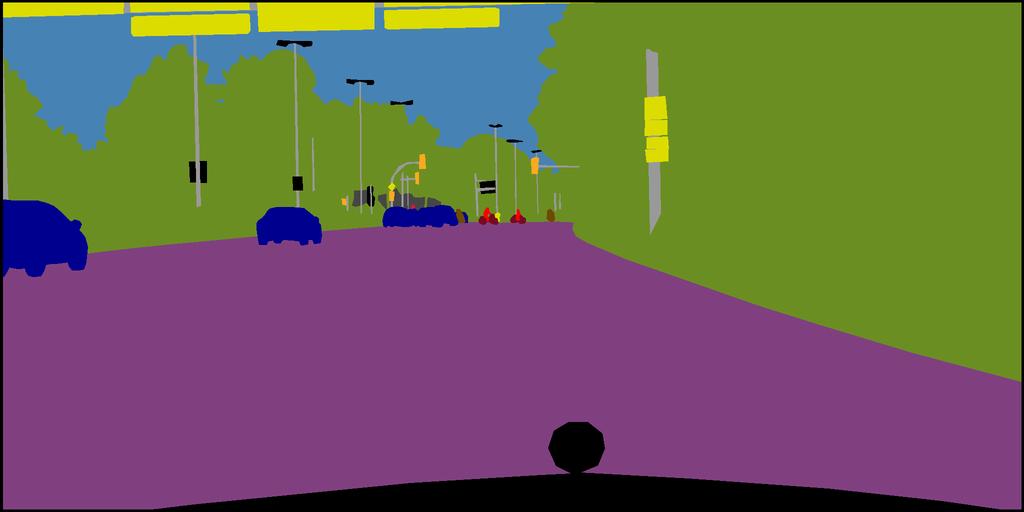  }\\
\hline

\verticaltext[22.5pt]{FedDyn(0.005)} &
\includegraphics[width=0.187\linewidth, height=0.10\linewidth]{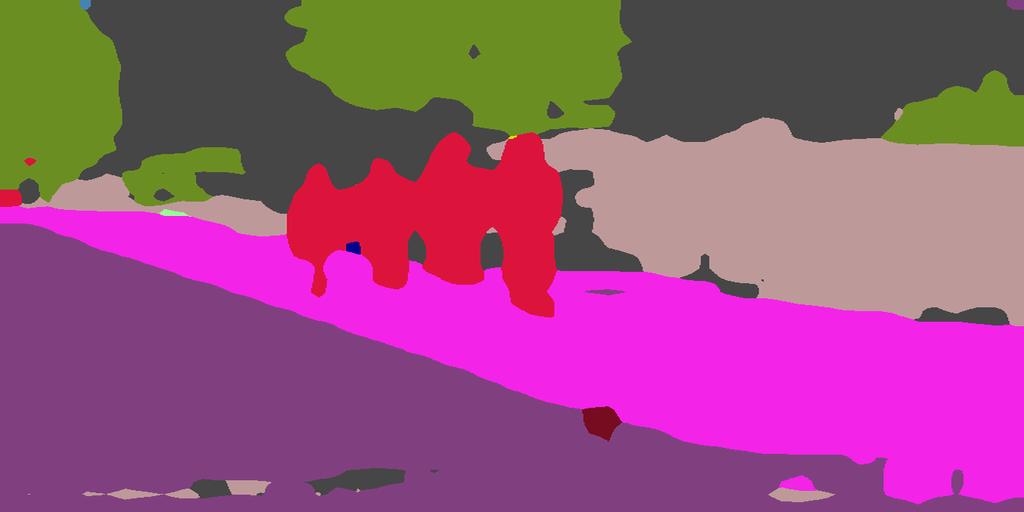} &\hspace{-0.47cm}
\includegraphics[width=0.187\linewidth, height=0.10\linewidth]{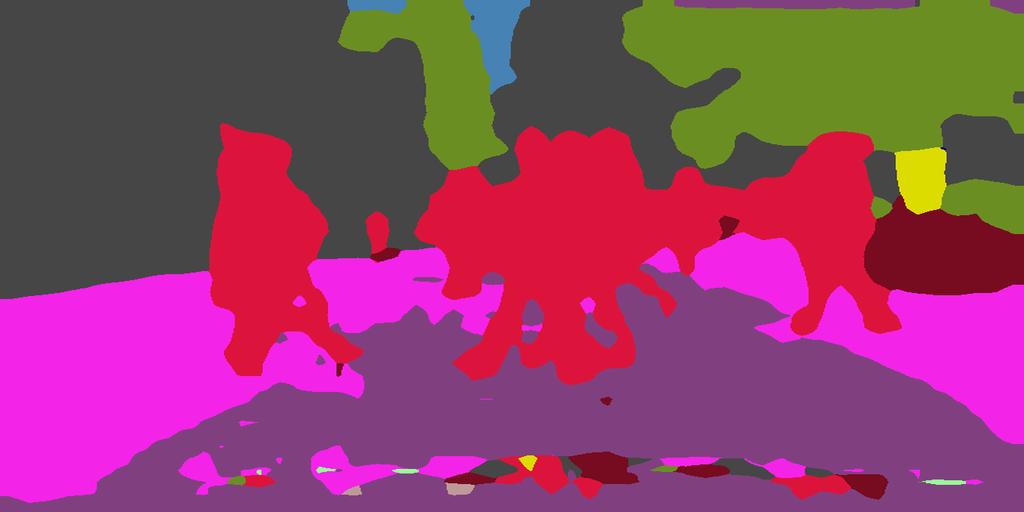} &\hspace{-0.47cm}
\includegraphics[width=0.187\linewidth, height=0.10\linewidth]{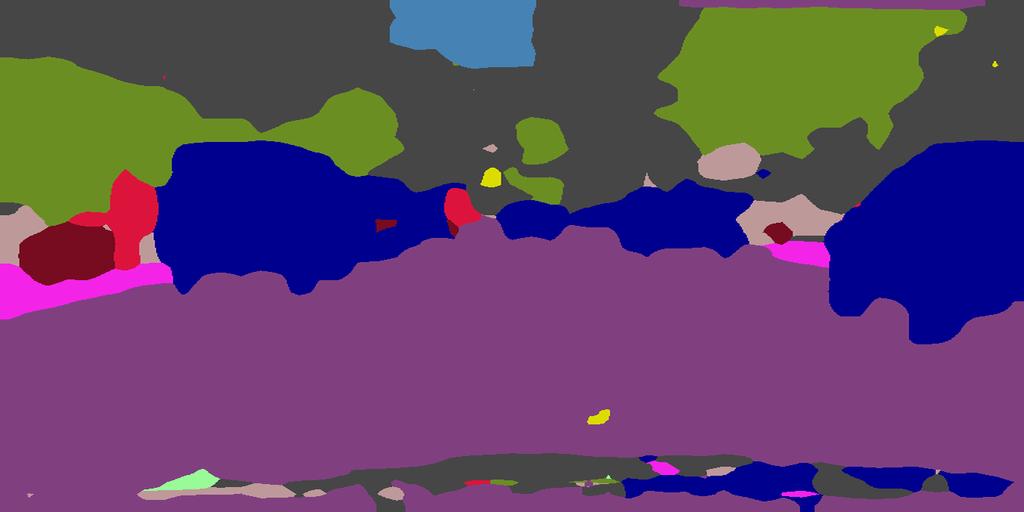} &\hspace{-0.47cm}
\includegraphics[width=0.187\linewidth, height=0.10\linewidth]{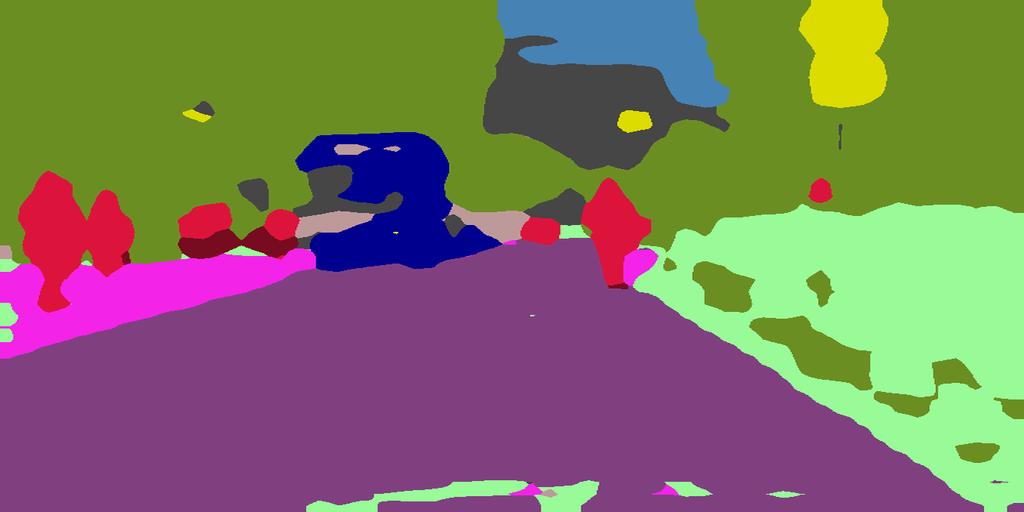  } &\hspace{-0.47cm}
\includegraphics[width=0.187\linewidth, height=0.10\linewidth]{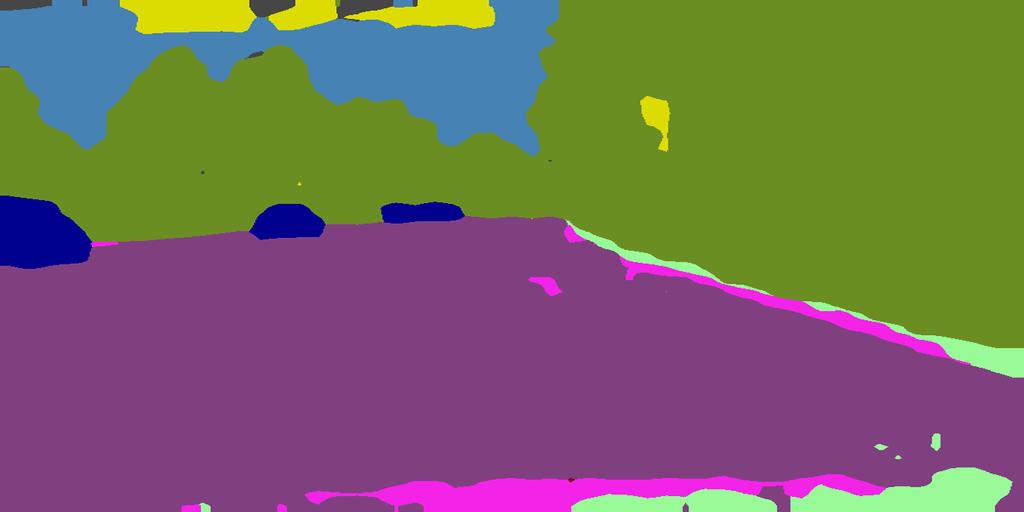  }\\
\hline

\verticaltext[22.5pt]{FedDyn(0.01)} &
\includegraphics[width=0.187\linewidth, height=0.10\linewidth]{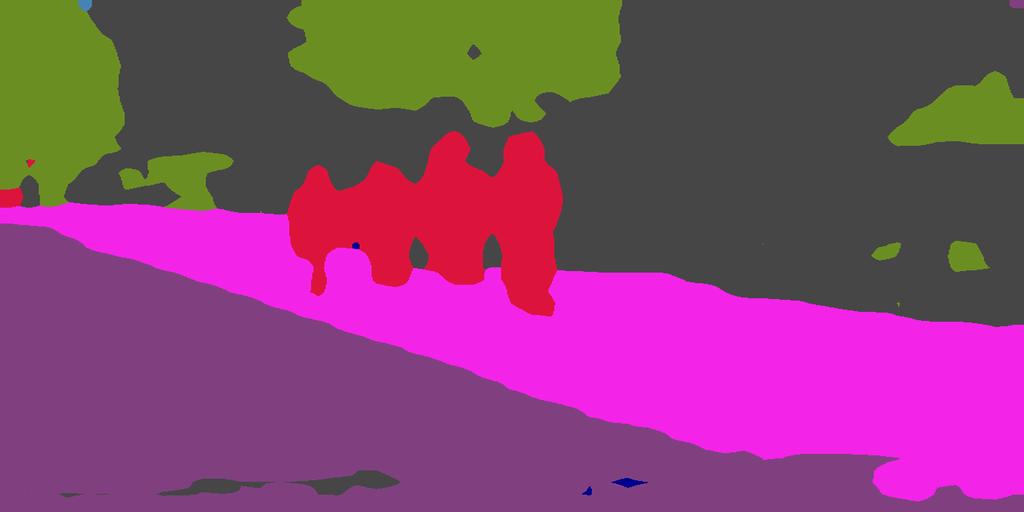} &\hspace{-0.47cm}
\includegraphics[width=0.187\linewidth, height=0.10\linewidth]{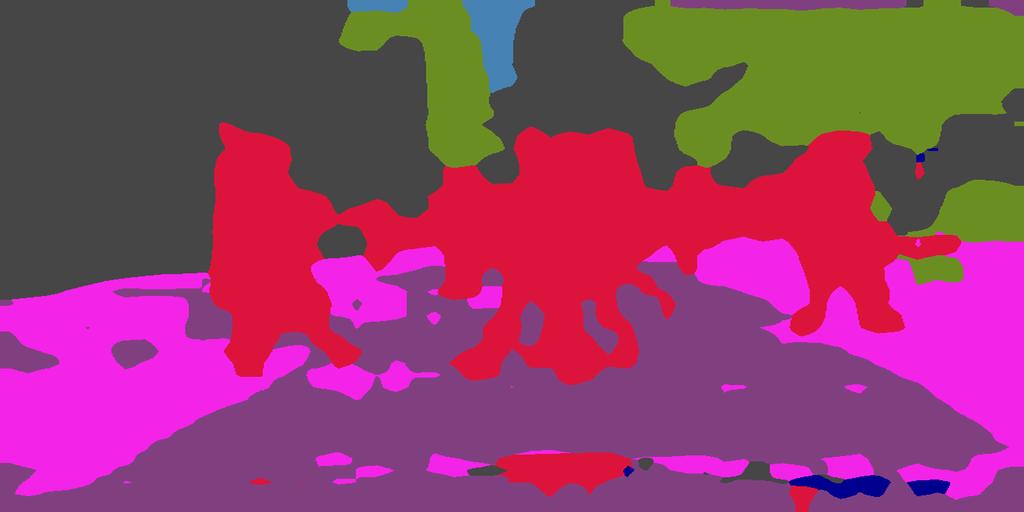} &\hspace{-0.47cm}
\includegraphics[width=0.187\linewidth, height=0.10\linewidth]{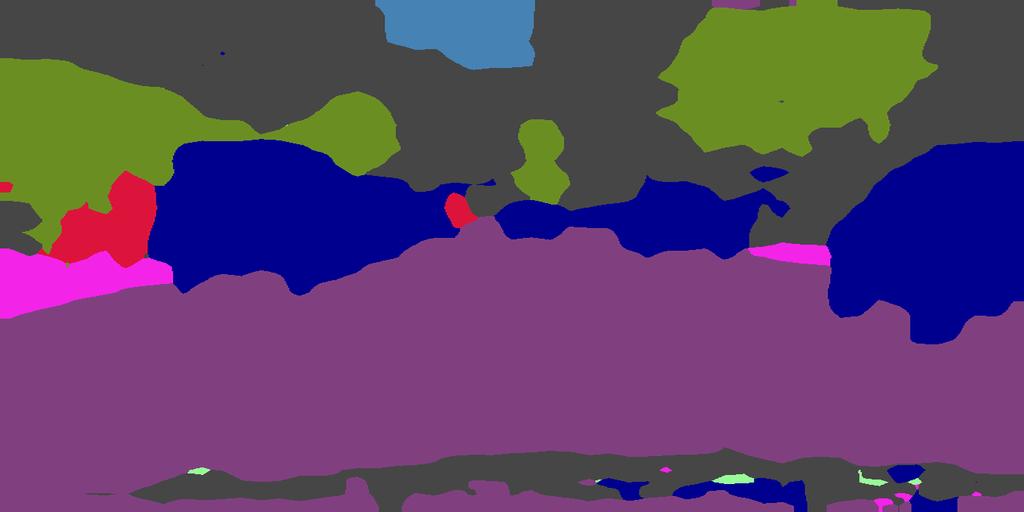} &\hspace{-0.47cm}
\includegraphics[width=0.187\linewidth, height=0.10\linewidth]{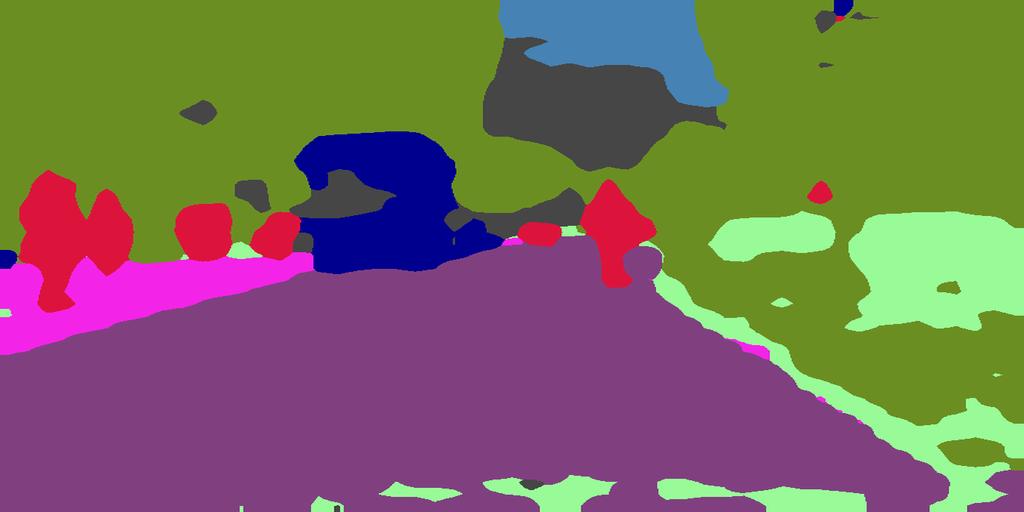  } &\hspace{-0.47cm}
\includegraphics[width=0.187\linewidth, height=0.10\linewidth]{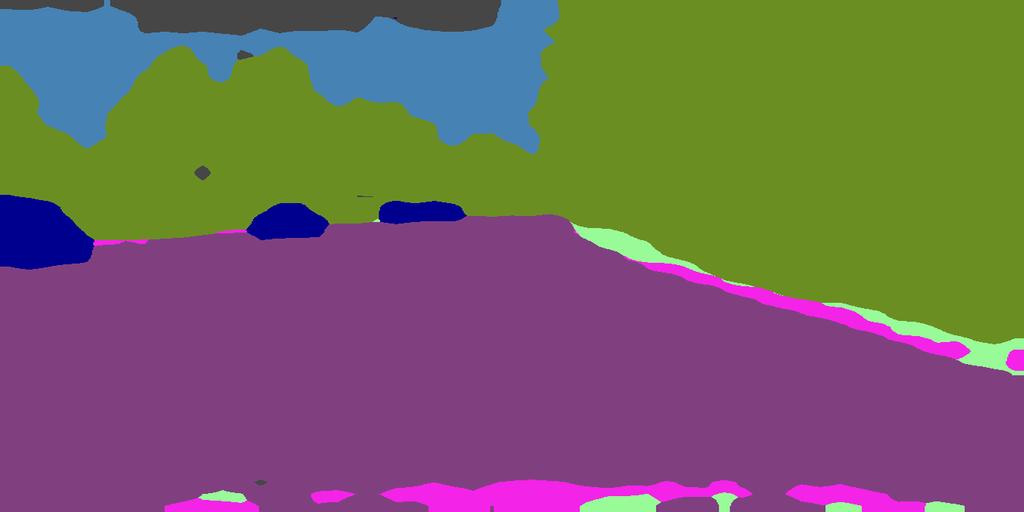  }\\
\hline

\verticaltext[22.5pt]{FedIR} &
\includegraphics[width=0.187\linewidth, height=0.10\linewidth]{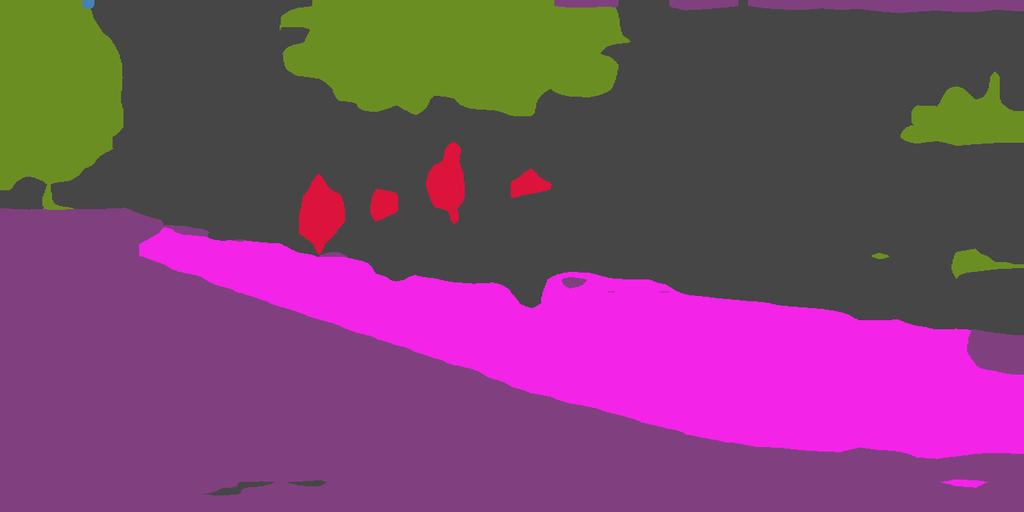} &\hspace{-0.47cm}
\includegraphics[width=0.187\linewidth, height=0.10\linewidth]{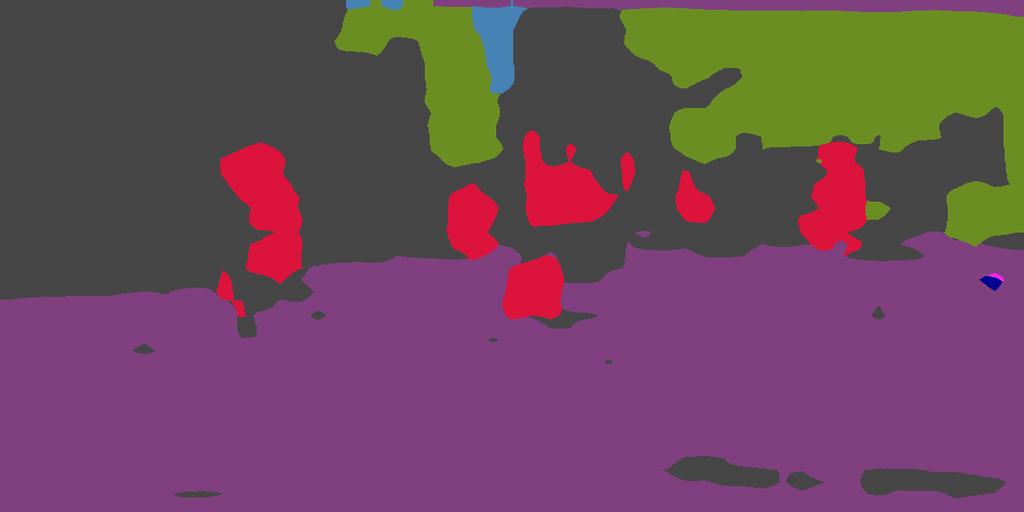} &\hspace{-0.47cm}
\includegraphics[width=0.187\linewidth, height=0.10\linewidth]{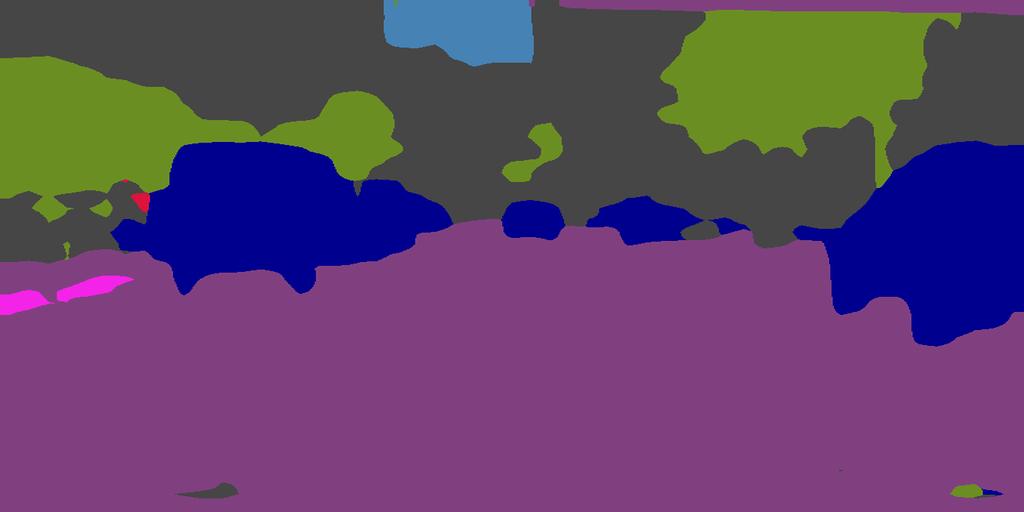} &\hspace{-0.47cm}
\includegraphics[width=0.187\linewidth, height=0.10\linewidth]{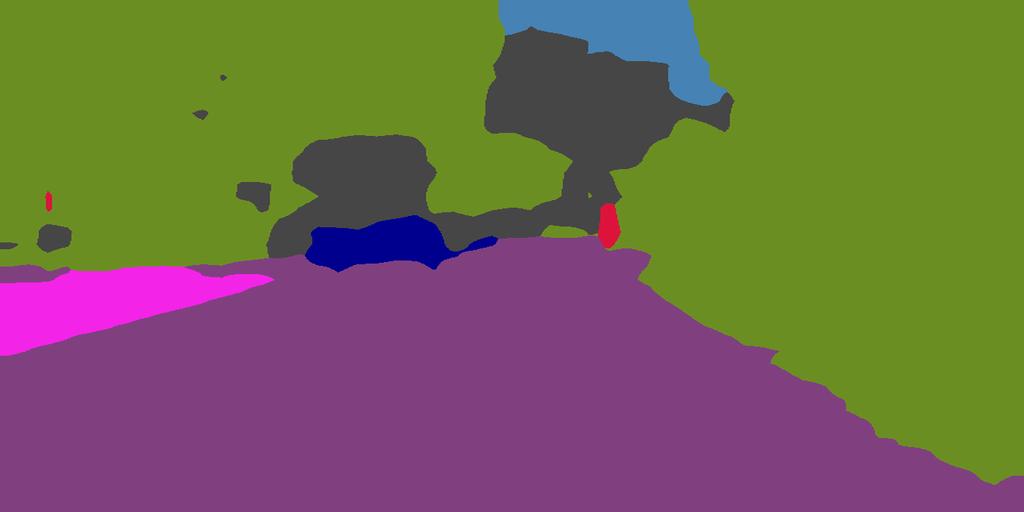  } &\hspace{-0.47cm}
\includegraphics[width=0.187\linewidth, height=0.10\linewidth]{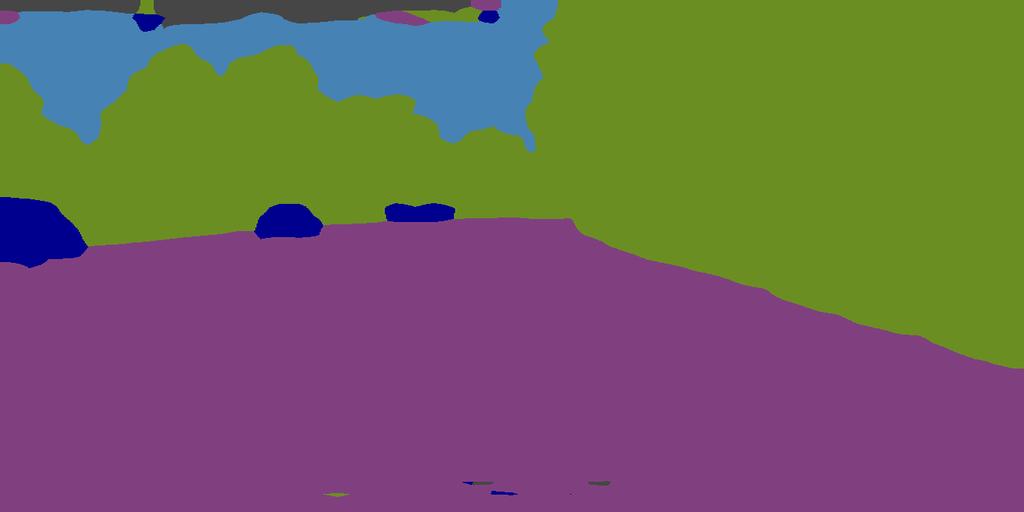  }\\
\hline

\verticaltext[22.5pt]{FedProx(0.005)} &
\includegraphics[width=0.187\linewidth, height=0.10\linewidth]{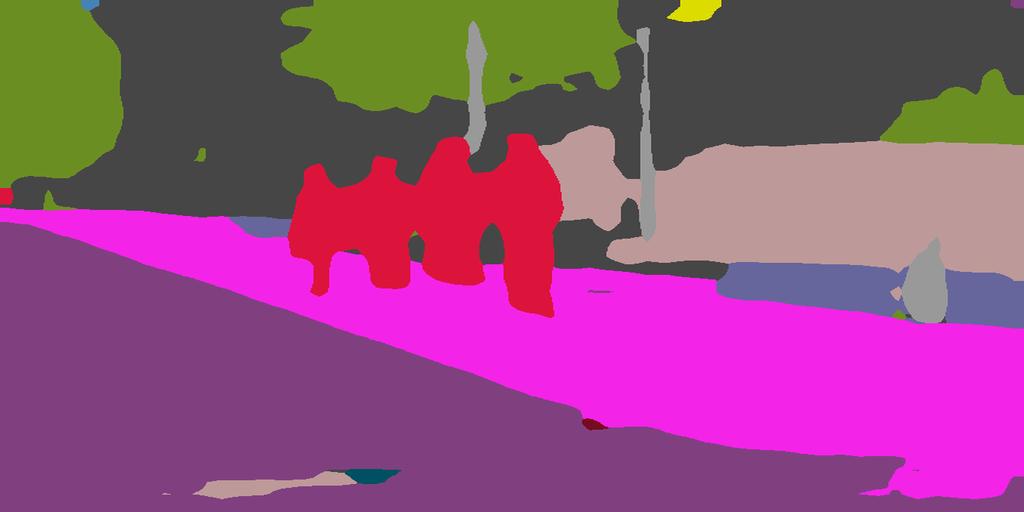} &\hspace{-0.47cm}
\includegraphics[width=0.187\linewidth, height=0.10\linewidth]{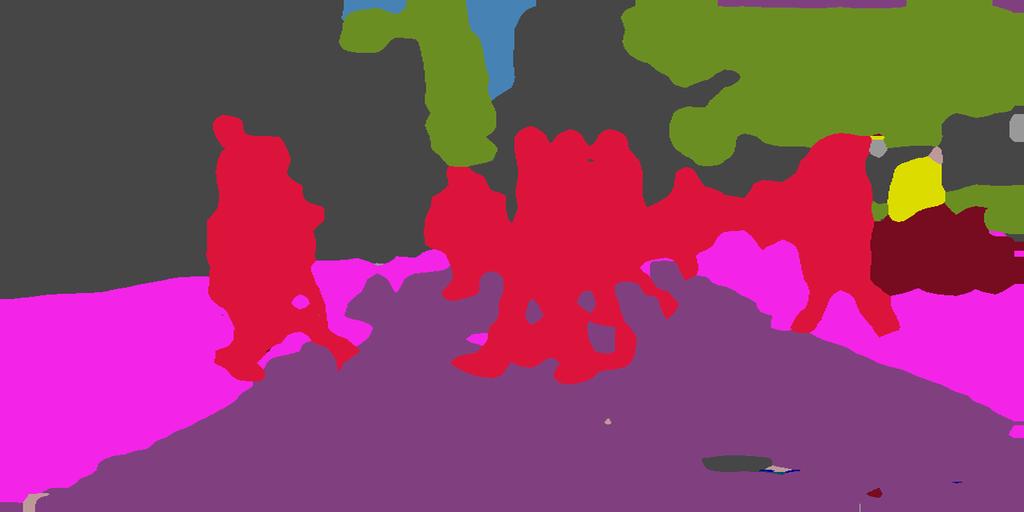} &\hspace{-0.47cm}
\includegraphics[width=0.187\linewidth, height=0.10\linewidth]{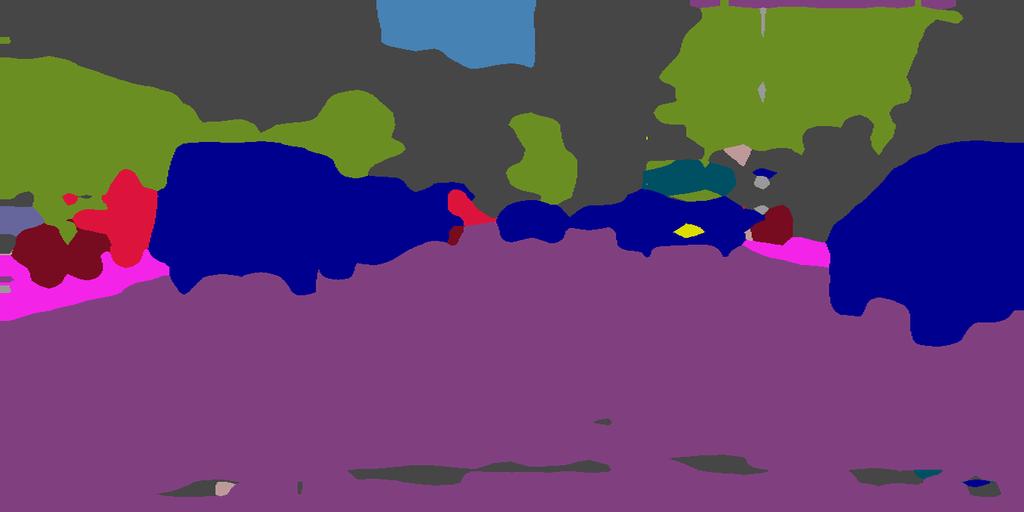} &\hspace{-0.47cm}
\includegraphics[width=0.187\linewidth, height=0.10\linewidth]{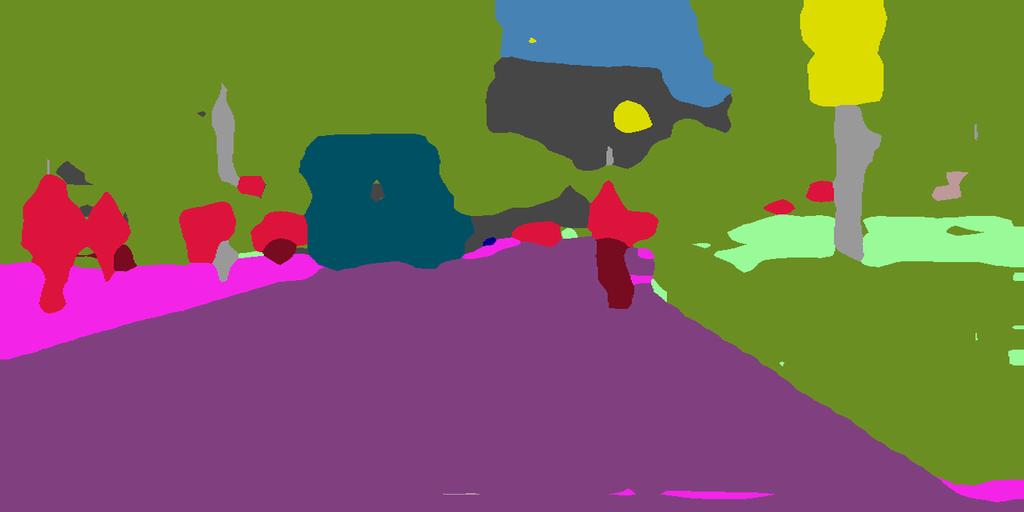  } &\hspace{-0.47cm}
\includegraphics[width=0.187\linewidth, height=0.10\linewidth]{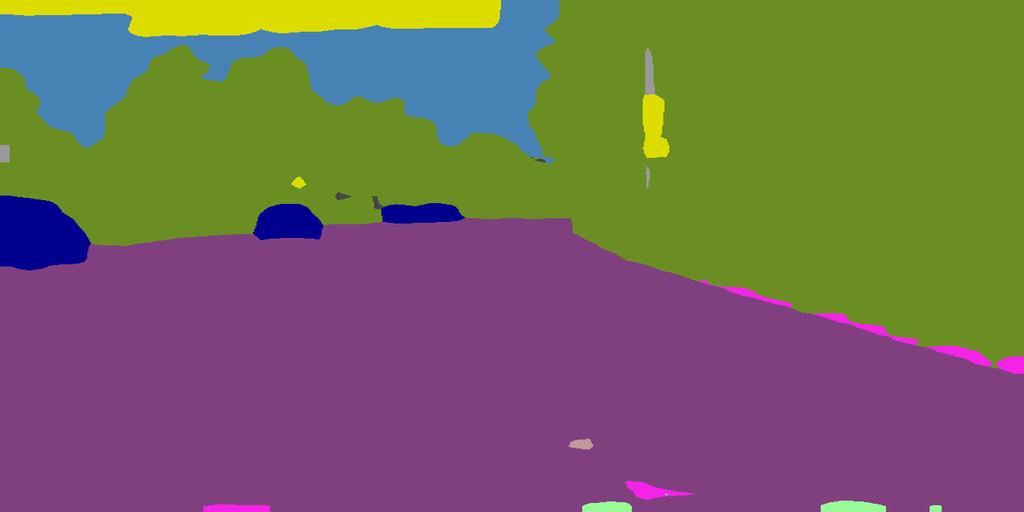  }\\
\hline

\verticaltext[22.5pt]{FedProx(0.01)} &
\includegraphics[width=0.187\linewidth, height=0.10\linewidth]{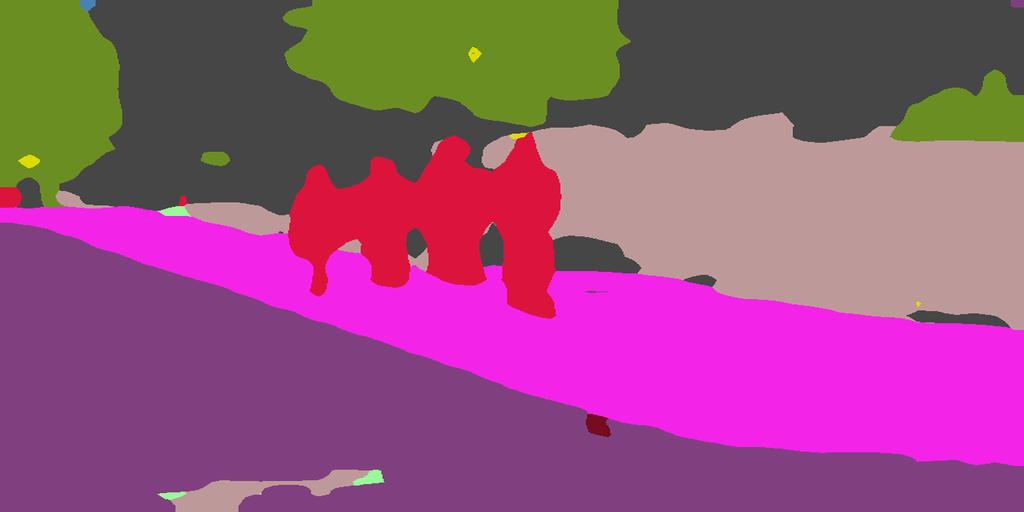} &\hspace{-0.47cm}
\includegraphics[width=0.187\linewidth, height=0.10\linewidth]{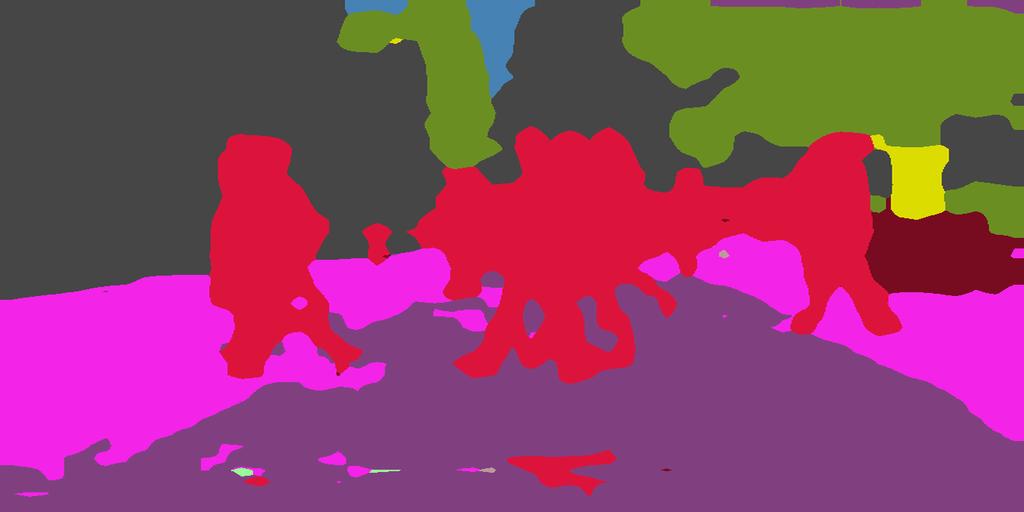} &\hspace{-0.47cm}
\includegraphics[width=0.187\linewidth, height=0.10\linewidth]{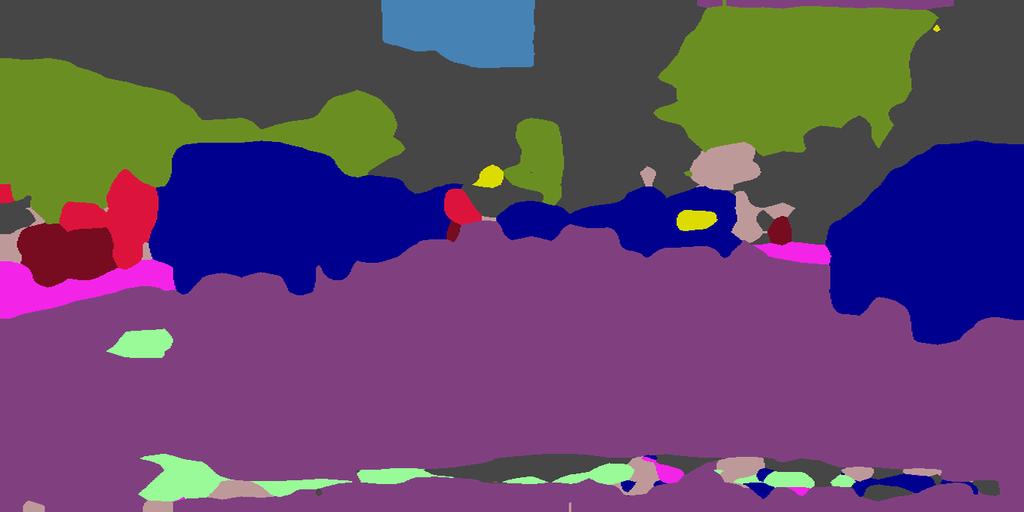} &\hspace{-0.47cm}
\includegraphics[width=0.187\linewidth, height=0.10\linewidth]{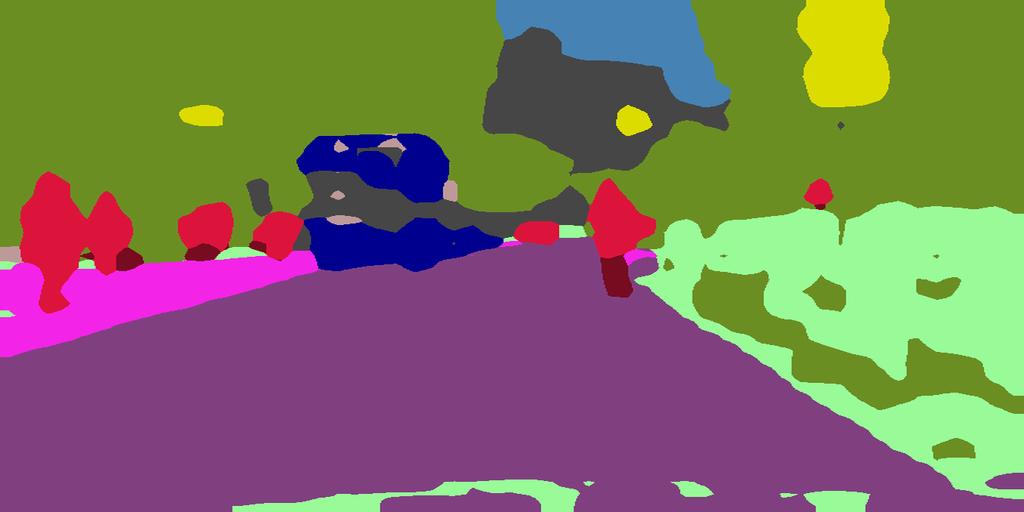  } &\hspace{-0.47cm}
\includegraphics[width=0.187\linewidth, height=0.10\linewidth]{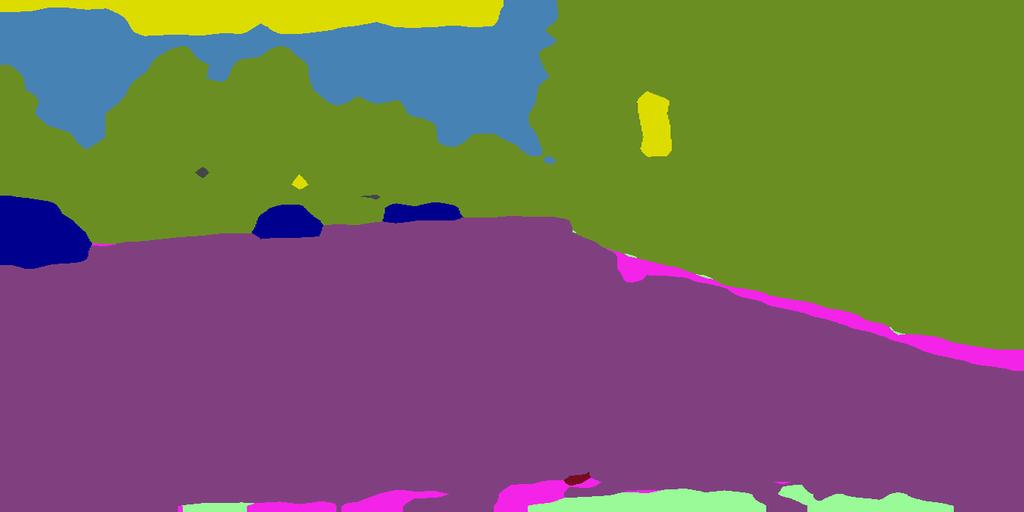  }\\
\hline

\verticaltext[23pt]{MOON} &
\includegraphics[width=0.187\linewidth, height=0.10\linewidth]{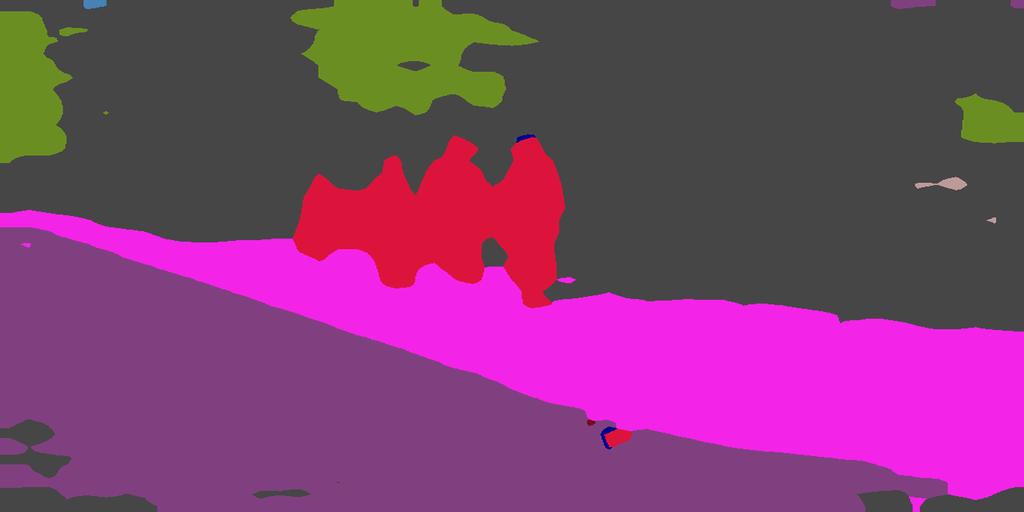} &\hspace{-0.47cm}
\includegraphics[width=0.187\linewidth, height=0.10\linewidth]{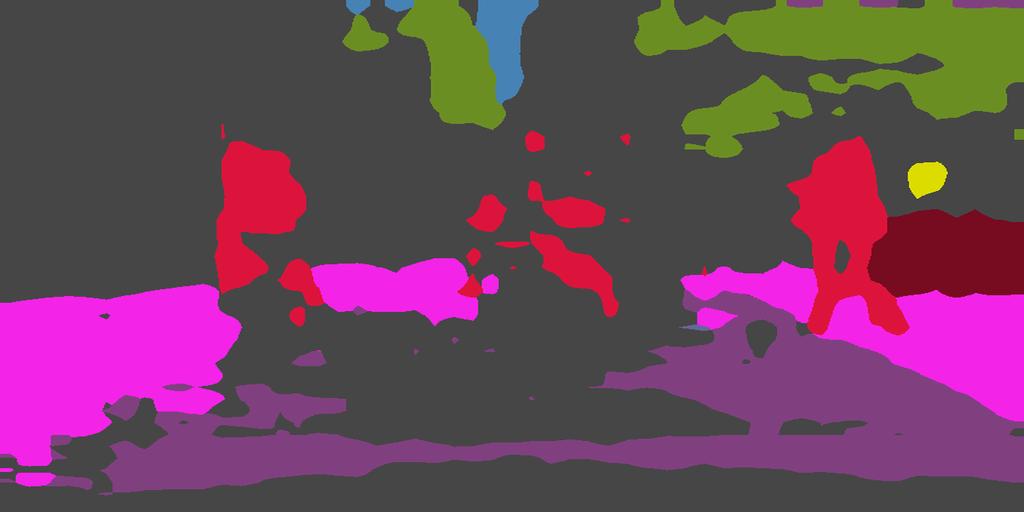} &\hspace{-0.47cm}
\includegraphics[width=0.187\linewidth, height=0.10\linewidth]{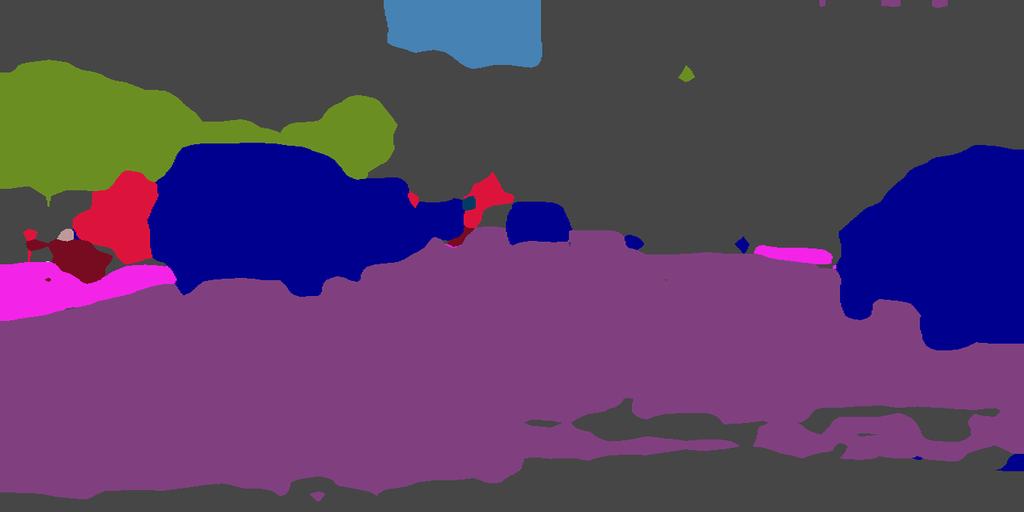} &\hspace{-0.47cm}
\includegraphics[width=0.187\linewidth, height=0.10\linewidth]{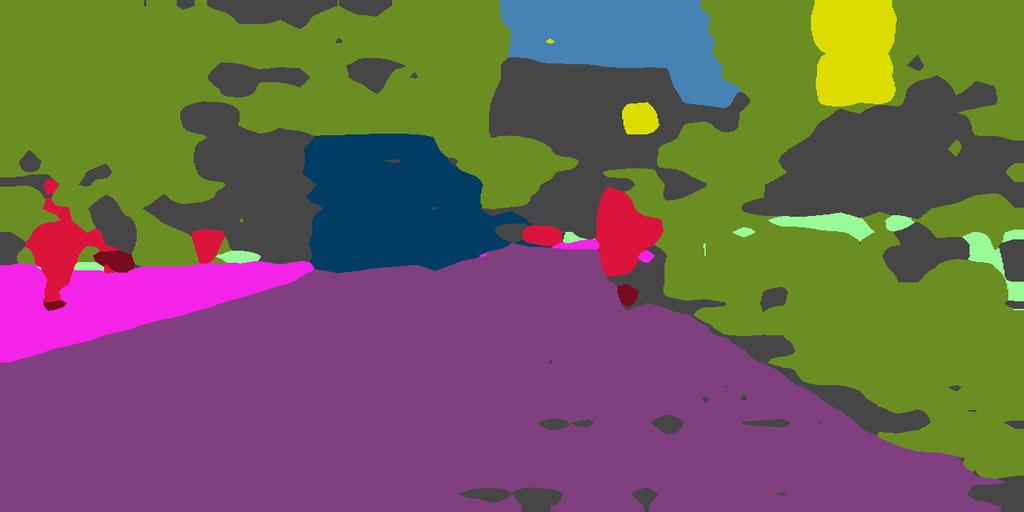  } &\hspace{-0.47cm}
\includegraphics[width=0.187\linewidth, height=0.10\linewidth]{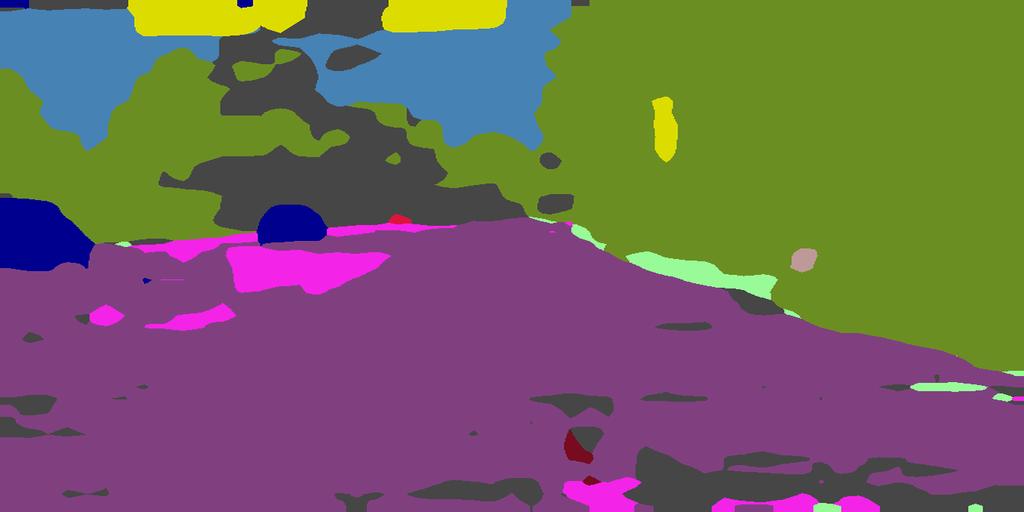  }\\
\hline

\verticaltext[22.5pt]{\textbf{FedEMA(ours)}} &
\includegraphics[width=0.187\linewidth, height=0.10\linewidth]{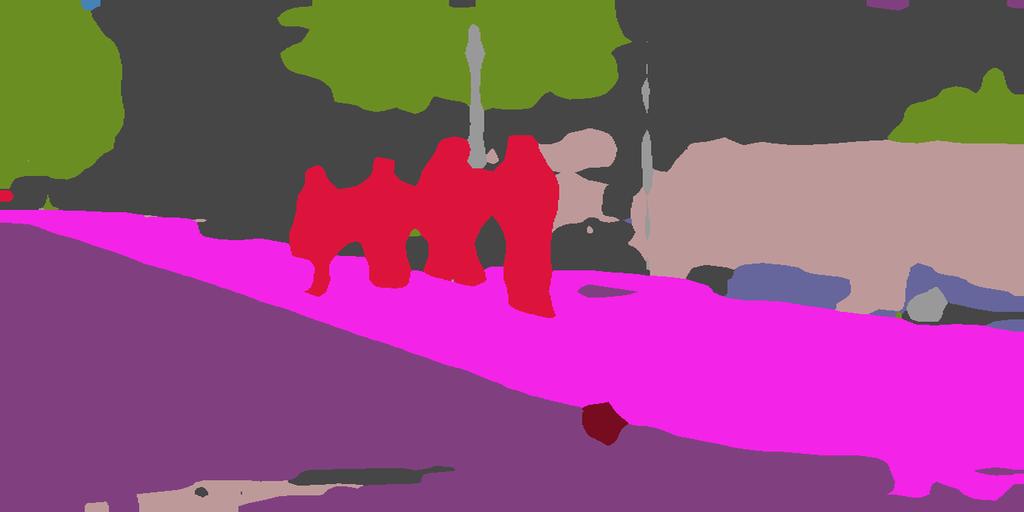} &\hspace{-0.47cm}
\includegraphics[width=0.187\linewidth, height=0.10\linewidth]{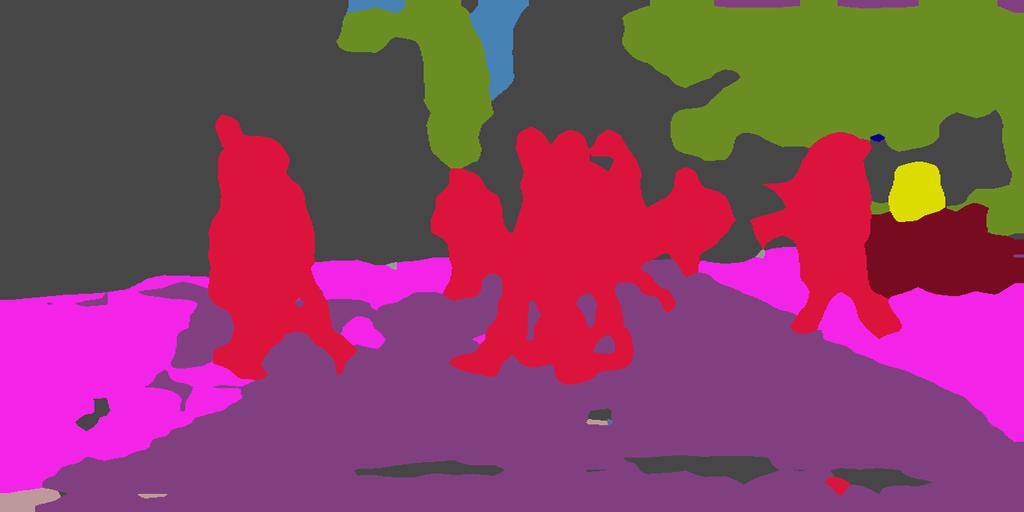} &\hspace{-0.47cm}
\includegraphics[width=0.187\linewidth, height=0.10\linewidth]{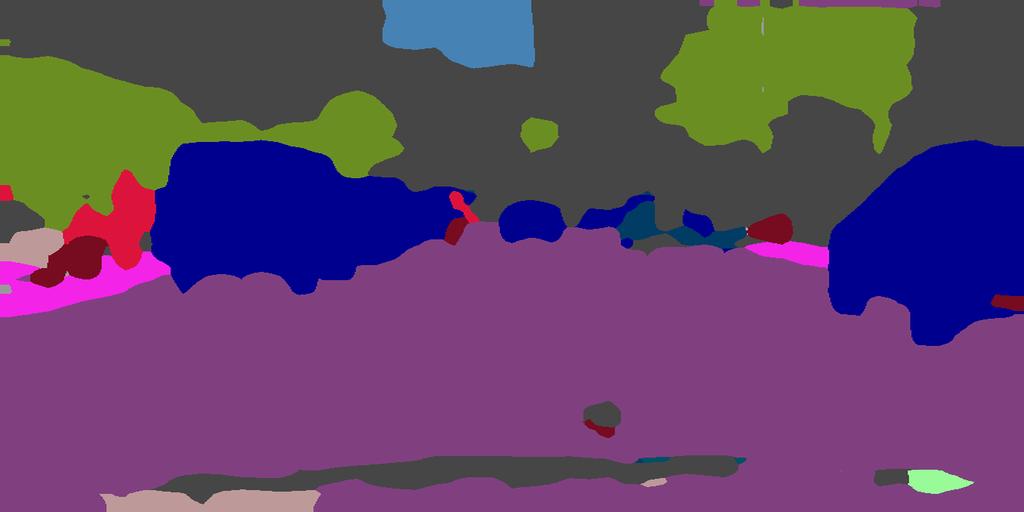} &\hspace{-0.47cm}
\includegraphics[width=0.187\linewidth, height=0.10\linewidth]{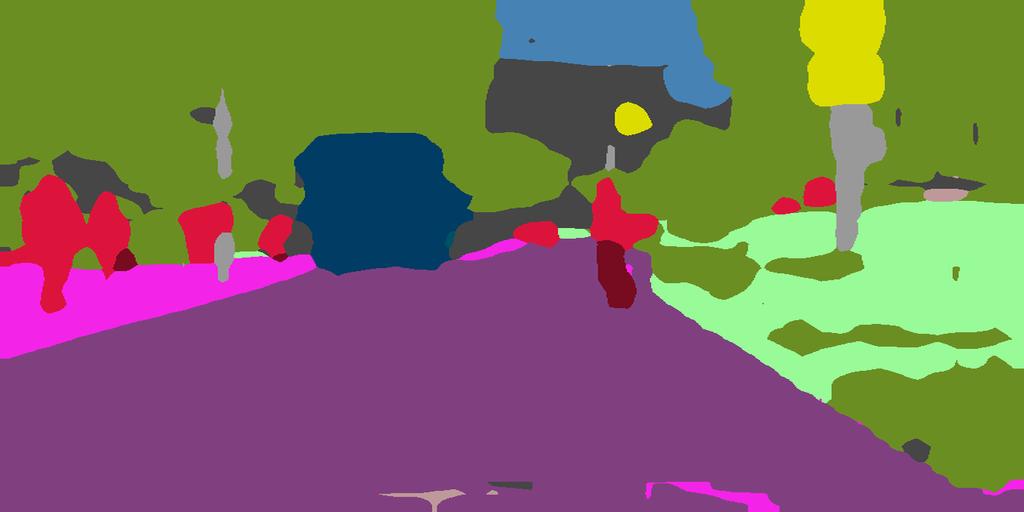  } &\hspace{-0.47cm}
\includegraphics[width=0.187\linewidth, height=0.10\linewidth]{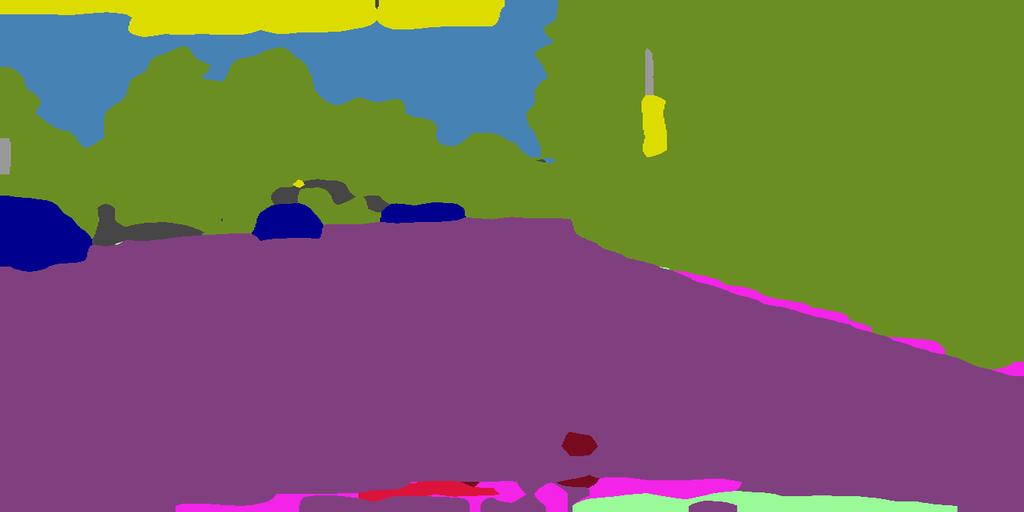  }\\ \hline
\end{tabularx}
\label{tab:semantic_pred}
\vspace{-0.5cm}
\end{table*}

In the quantitatively analysis, we conduct a set of experiments to compare the performance of multiple FL algorithms on CNN-based DeepLabv3+ model \cite{chen2018encoderdecoder} and Transformer-based TopFormer model \cite{zhang2022topformer}. Their results are presented in \Cref{Tab:metrics_deeplabv3} and \Cref{Tab:metrics_topformer}, respectively. Notably, in these two tables, the highest performance metric is highlighted in bold, and the second highest is underlined. From \Cref{Tab:metrics_deeplabv3} and \Cref{Tab:metrics_topformer}, we can observe following patterns: (I) For both datasets and both models, the proposed FedEMA achieves almost the best performance. Taking the combination of Cityscapes dataset and DeepLabv3+ model as example, FedEMA outperforms the second-best FL algorithm (\ie, FedProx (0.005)) by margins of (60.62 - 56.59) / 56.59  = 7.12\%, (51.60 - 48.41) / 48.41  = 6.59\%, (64.33 - 58.41) / 58.41  = 10.14\%, and (59.17 - 56.78) / 56.78  = 4.21\% in mIoU, mF1, mPrecision, and mRecall, respectively. (II) Upon inspecting \Cref{Tab:metrics_deeplabv3} and \Cref{Tab:metrics_topformer}, we can observe that a negative correlation between the performance of FL algorithms and dataset complexity. Specifically, the majority of FL algorithms tend to underperform on the complex Cityscapes dataset relative to their performance on the simpler CamVid dataset. (III) The model architecture has a substantial impact on the performance of FL algorithms and even on the convergence of FL algorithms. For example, MOON achieves relatively good performance on both Cityscapes dataset and CamVid dataset for DeepLabv3+ model, whereas MOON diverges on both datasets for TopFormer model, across all adopted metrics. \Cref{Fig.Metrics} visually confirms above insights. It is worth to note that unlike other FL algorithms that generally converge after a few training rounds, the proposed FedEMA continues to evolve, adapting to dynamic training data throughout the training process. 

\subsubsection{Qualitative Performance Comparison}
In the qualitative comparison, \Cref{tab:semantic_pred} compares multiple FL algorithms, including FedDyn(0.005), FedDyn(0.01), FedIR, FedProx(0.005), FedProx(0.01), MOON, along with our proposed FedEMA, on five RGB images from diverse AD scenarios. To assess the predictive accuracy of each FL algorithm, we evaluated how close their outputs match the ground truth and the original images. Our comparison shows that FedEMA delivers more precise predictions of both the overall scenes and detailed aspects across all images.

\begin{figure}[tp]
\centering
\subfloat[\footnotesize mIoU]{\includegraphics[width=0.48\linewidth]{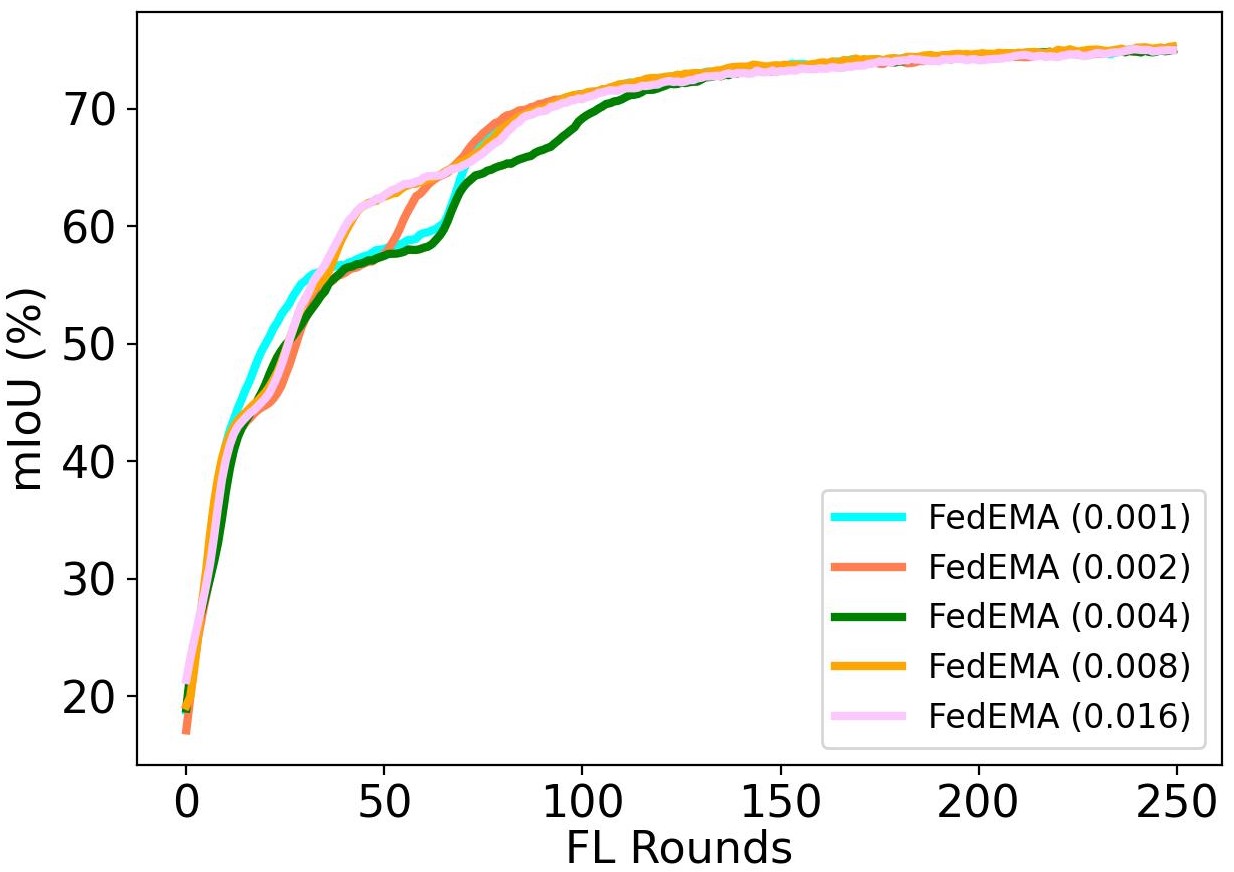}
\label{Fig.Metrics_mIoU}
}
\subfloat[\footnotesize mF1]{\includegraphics[width=0.48\linewidth]{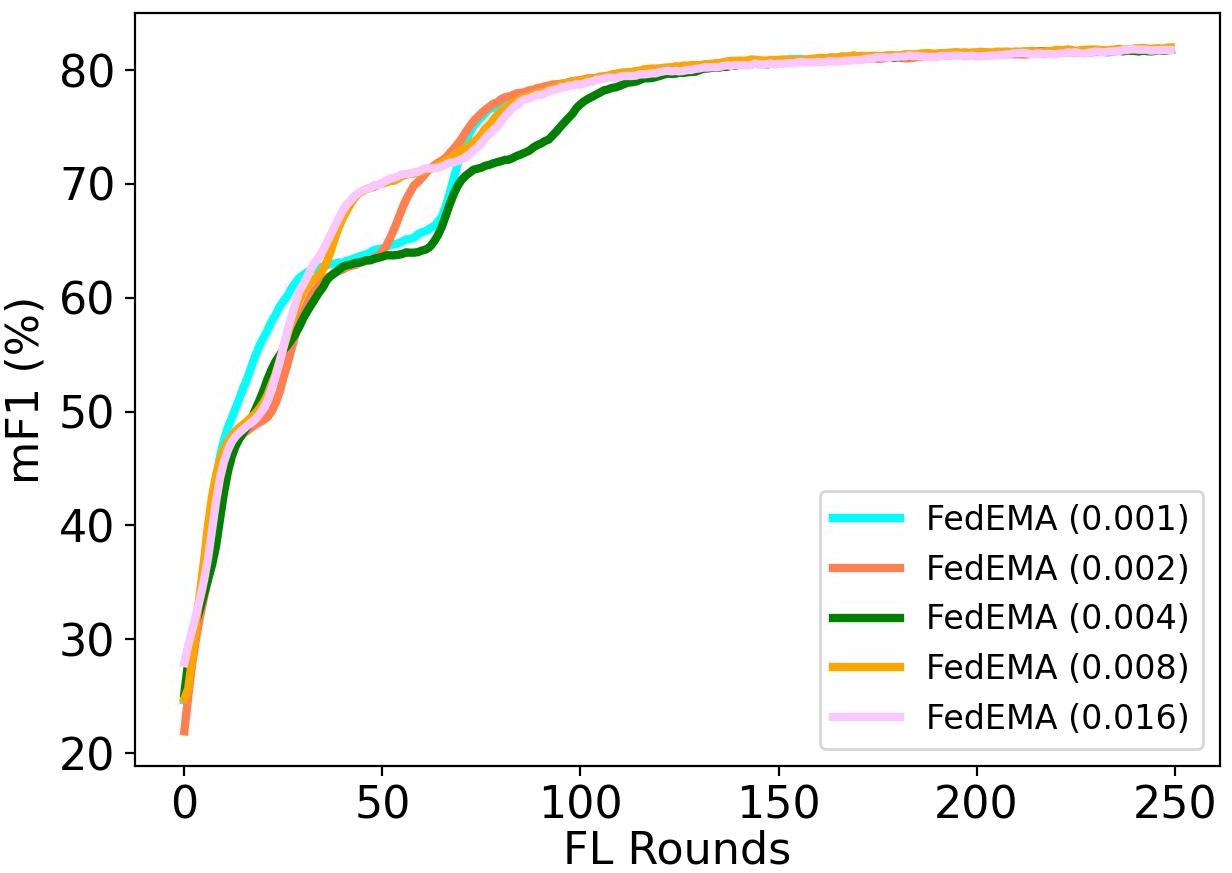}
\label{Fig.Metrics_mF1}
}
\vspace{-0.15cm}
\caption{Performance and convergence comparison with different regularization coefficient $\lambda$ while keeping window size $N=5$.}
\label{Fig.Metrics_abl_coef}
\vspace{0.2cm}
\end{figure}

\begin{figure}[tp]
\centering
\subfloat[\footnotesize mIoU]{\includegraphics[width=0.48\linewidth]{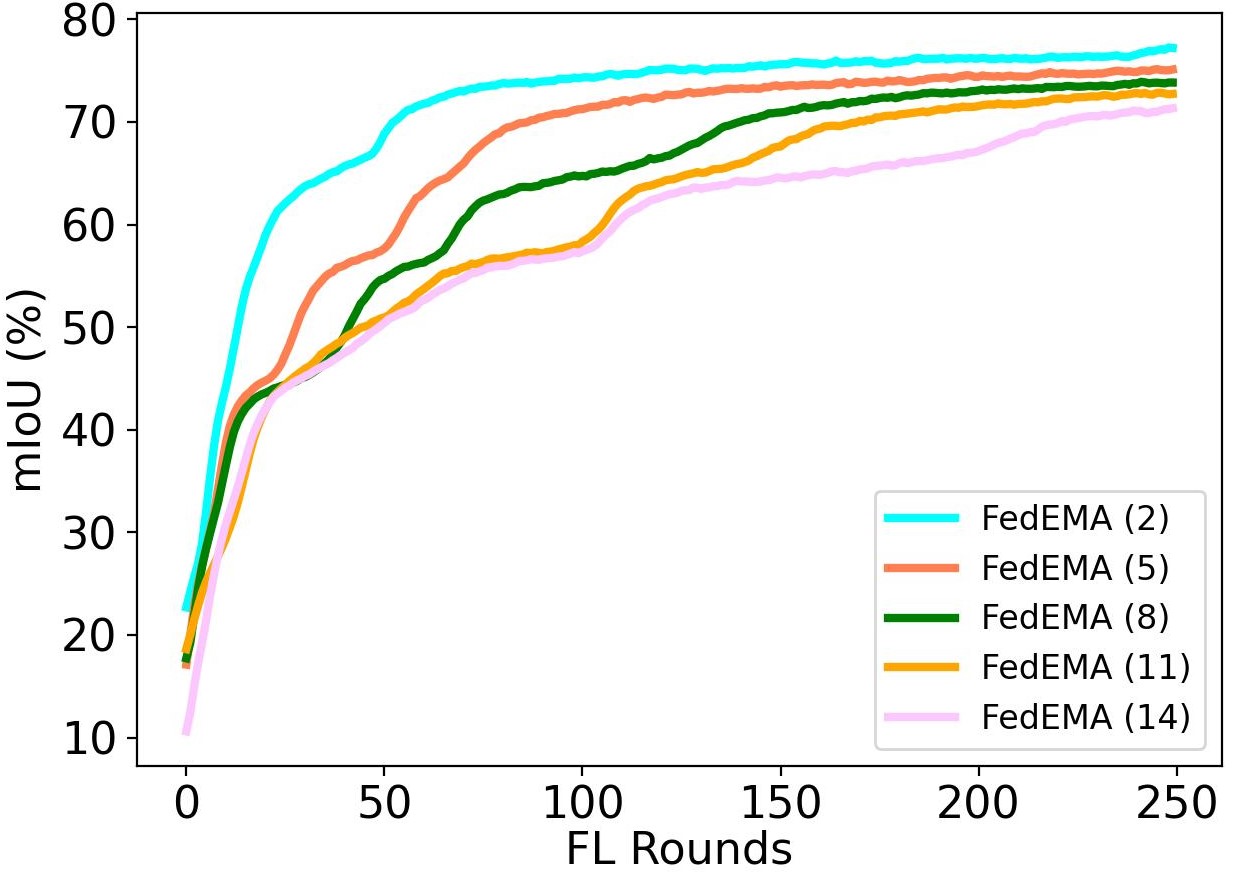}
\label{Fig.Metrics_mIoU}
}
\subfloat[\footnotesize mF1]{\includegraphics[width=0.48\linewidth]{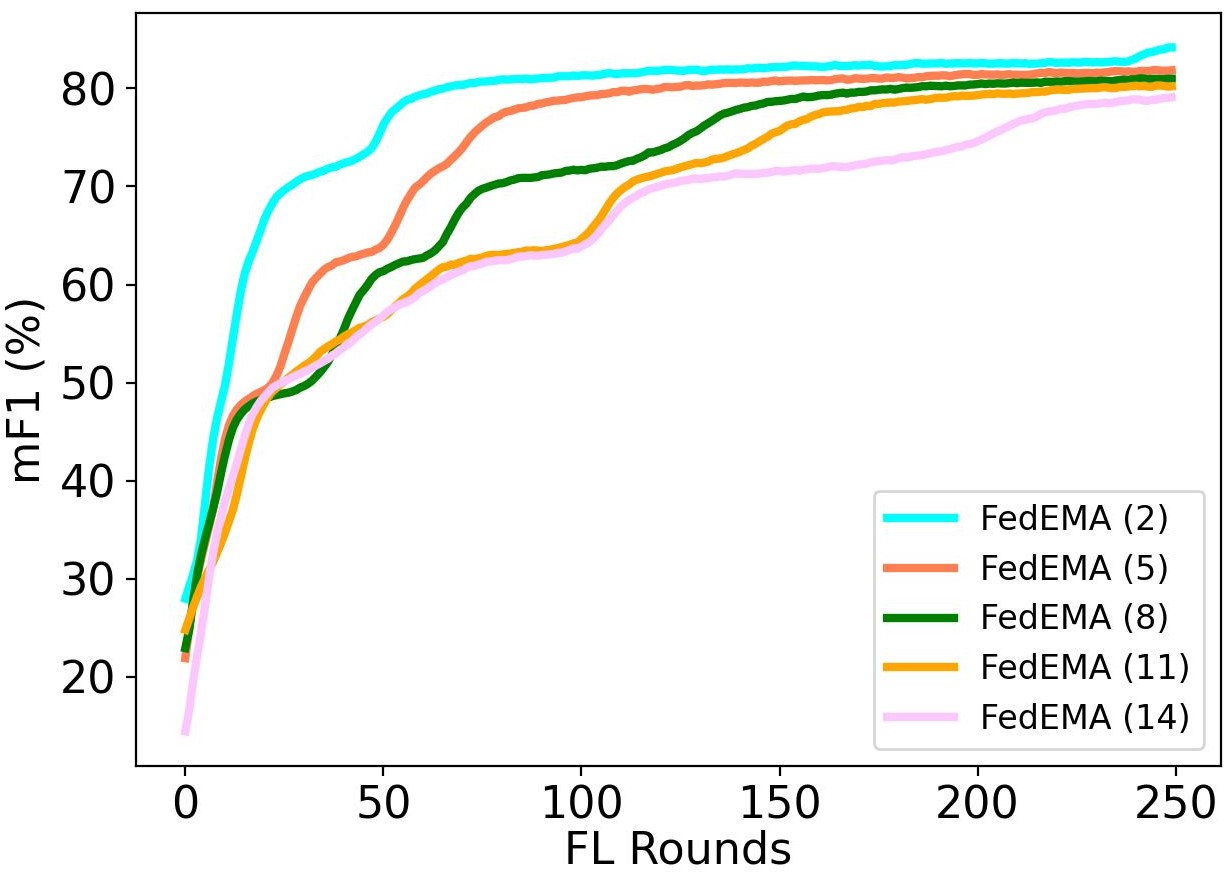}
\label{Fig.Metrics_mF1}
}
\vspace{-0.15cm}
\caption{Performance and convergence comparison with different window size $N$ while keeping regularization coefficient $\lambda=0.002$.}
\label{Fig.Metrics_abl_ws}
\vspace{0.1cm}
\end{figure}

\subsection{Ablation Study}
In this part, we conduct two aspects of ablation studies for the proposed FedEMA method. On the one hand, we conduct experiments to explore how negative entropy regularization coefficient $\lambda$ affects the performance and convergence of the proposed FedEMA algorithm. The results can be found in \Cref{Fig.Metrics_abl_coef}. From \Cref{Fig.Metrics_abl_coef}, We can conclude that the coefficient $\lambda$ significantly impacts the convergence of the FL model, yet it has minimal influence on the final performance, where a higher $\lambda$ results in quicker convergence. This result aligns with the convergence analysis in \Cref{FedEMA_convergence}. On the other hand, we carry out a set of experiments to investigate the impact of window size $N$ on the performance and convergence of the proposed FedEMA. The results are illustrated in \Cref{Fig.Metrics_abl_ws}. From \Cref{Fig.Metrics_abl_ws}, we can observe that the window size significantly affects both the performance and convergence of the FL model. Specifically, a smaller window size leads to faster convergence and improved performance.

\section{Conclusion}
In this study, we attempt to improve S3U model generalization in FedAD systems. FedEMA was proposed to solve the catastrophic forgetting problem in evolving driving environments. We conduct comprehensive experiments and compare the results with current SOTA approaches. The findings reveal that FedEMA can improve S3U model generalization by overcoming the catastrophic forgetting problem. Future work includes applying FedEMA to a wider range of AD tasks and integrating multi-modal data into FedEMA.


\begin{thebibliography}{10}

\bibitem{10416354}
J.~Rückin, F.~Magistri, C.~Stachniss, and M.~Popović, ``Semi-supervised active learning for semantic segmentation in unknown environments using informative path planning,'' \emph{IEEE Robotics and Automation Letters}, vol.~9, no.~3, pp. 2662--2669, 2024.

\bibitem{10160999}
N.~Kim, T.~Son, J.~Pahk, C.~Lan, W.~Zeng, and S.~Kwak, ``Wedge: Web-image assisted domain generalization for semantic segmentation,'' in \emph{2023 IEEE International Conference on Robotics and Automation (ICRA)}, 2023, pp. 9281--9288.

\bibitem{10049523}
Z.~Feng, Y.~Guo, and Y.~Sun, ``Cekd: Cross-modal edge-privileged knowledge distillation for semantic scene understanding using only thermal images,'' \emph{IEEE Robotics and Automation Letters}, vol.~8, no.~4, pp. 2205--2212, 2023.

\bibitem{kou2025enhancing}
W.-B. Kou, Q.~Lin, M.~Tang, S.~Wang, R.~Ye, G.~Zhu, and Y.-C. Wu, ``Enhancing large vision model in street scene semantic understanding through leveraging posterior optimization trajectory,'' \emph{arXiv preprint arXiv:2501.01710}, 2025.

\bibitem{10342110}
H.~Son and J.~Weiland, ``Lightweight semantic segmentation network for semantic scene understanding on low-compute devices,'' in \emph{2023 IEEE/RSJ International Conference on Intelligent Robots and Systems (IROS)}, 2023, pp. 62--69.

\bibitem{10342254}
Z.~Chen, Z.~Ding, J.~M. Gregory, and L.~Liu, ``Ida: Informed domain adaptive semantic segmentation,'' in \emph{2023 IEEE/RSJ International Conference on Intelligent Robots and Systems(IROS)}, 2023, pp. 90--97.

\bibitem{10341639}
J.~Li, W.~Shi, D.~Zhu, G.~Zhang, X.~Zhang, and J.~Li, ``Featdanet: Feature-level domain adaptation network for semantic segmentation,'' in \emph{2023 IEEE/RSJ International Conference on Intelligent Robots and Systems (IROS)}, 2023, pp. 3873--3880.

\bibitem{muhammad2022vision}
K.~Muhammad, T.~Hussain, H.~Ullah, J.~Del~Ser, M.~Rezaei, N.~Kumar, M.~Hijji, P.~Bellavista, and V.~H.~C. de~Albuquerque, ``Vision-based semantic segmentation in scene understanding for autonomous driving: Recent achievements, challenges, and outlooks,'' \emph{IEEE Transactions on Intelligent Transportation Systems}, 2022.

\bibitem{https://doi.org/10.48550/arxiv.1602.05629}
H.~B. McMahan, E.~Moore, D.~Ramage, S.~Hampson, and B.~A.~y. Arcas, ``Communication-efficient learning of deep networks from decentralized data,'' 2016.

\bibitem{10342134}
W.-B. Kou, S.~Wang, G.~Zhu, B.~Luo, Y.~Chen, D.~W. Kwan~Ng, and Y.-C. Wu, ``Communication resources constrained hierarchical federated learning for end-to-end autonomous driving,'' in \emph{2023 IEEE/RSJ International Conference on Intelligent Robots and Systems (IROS)}, 2023, pp. 9383--9390.

\bibitem{wu2024hierarchical}
H.-T. Wu, H.~Li, H.-L. Chi, W.-B. Kou, Y.-C. Wu, and S.~Wang, ``A hierarchical federated learning framework for collaborative quality defect inspection in construction,'' \emph{Engineering Applications of Artificial Intelligence}, vol. 133, p. 108218, 2024.

\bibitem{kou2024pfedlvm}
W.-B. Kou, Q.~Lin, M.~Tang, S.~Xu, R.~Ye, Y.~Leng, S.~Wang, Z.~Chen, G.~Zhu, and Y.-C. Wu, ``pfedlvm: A large vision model (lvm)-driven and latent feature-based personalized federated learning framework in autonomous driving,'' \emph{arXiv preprint arXiv:2405.04146}, 2024.

\bibitem{kou2024fedrc}
W.-B. Kou, Q.~Lin, M.~Tang, S.~Wang, G.~Zhu, and Y.-C. Wu, ``Fedrc: A rapid-converged hierarchical federated learning framework in street scene semantic understanding,'' in \emph{2024 IEEE/RSJ International Conference on Intelligent Robots and Systems (IROS)}.\hskip 1em plus 0.5em minus 0.4em\relax IEEE, 2024, pp. 2578--2585.

\bibitem{kou2025fast}
W.-B. Kou, Q.~Lin, M.~Tang, R.~Ye, S.~Wang, G.~Zhu, and Y.-C. Wu, ``Fast-convergent and communication-alleviated heterogeneous hierarchical federated learning in autonomous driving,'' \emph{IEEE Transactions on Intelligent Transportation Systems}, 2025.

\bibitem{Cordts2016Cityscapes}
M.~Cordts, M.~Omran, S.~Ramos, T.~Rehfeld, M.~Enzweiler, R.~Benenson, U.~Franke, S.~Roth, and B.~Schiele, ``The cityscapes dataset for semantic urban scene understanding,'' in \emph{Proc. of the IEEE Conference on Computer Vision and Pattern Recognition (CVPR)}, 2016.

\bibitem{brostow2008segmentation}
G.~J. Brostow, J.~Shotton, J.~Fauqueur, and R.~Cipolla, ``Segmentation and recognition using structure from motion point clouds,'' in \emph{Proc. European Conference on Computer Vision of the (ECCV)}, 2008.

\bibitem{jiang2019multi}
Y.~Jiang, H.~Yedidsion, S.~Zhang, G.~Sharon, and P.~Stone, ``Multi-robot planning with conflicts and synergies,'' \emph{Autonomous Robots}, vol.~43, no.~8, pp. 2011--2032, 2019.

\bibitem{10494721}
M.~Kang, S.~Wang, S.~Zhou, K.~Ye, J.~Jiang, and N.~Zheng, ``Ffinet: Future feedback interaction network for motion forecasting,'' \emph{IEEE Transactions on Intelligent Transportation Systems}, pp. 1--12, 2024.

\bibitem{10802211}
M.~Herb, N.~Navab, and F.~Tombari, ``Neural semantic map-learning for autonomous vehicles,'' in \emph{2024 IEEE/RSJ International Conference on Intelligent Robots and Systems (IROS)}, 2024, pp. 1062--1069.

\bibitem{10372140}
J.~Wu, J.~Ruenz, H.~Berkemeyer, L.~Dixon, and M.~Althoff, ``Goal-oriented pedestrian motion prediction,'' \emph{IEEE Transactions on Intelligent Transportation Systems}, vol.~25, no.~6, pp. 5282--5298, 2024.

\bibitem{9165167}
Y.~Xiao, F.~Codevilla, A.~Gurram, O.~Urfalioglu, and A.~M. López, ``Multimodal end-to-end autonomous driving,'' \emph{IEEE Transactions on Intelligent Transportation Systems}, vol.~23, no.~1, pp. 537--547, 2022.

\bibitem{10414408}
W.~Zheng, X.~Jiang, Z.~Fang, and Y.~Gao, ``Tv-net: A structure-level feature fusion network based on tensor voting for road crack segmentation,'' \emph{IEEE Transactions on Intelligent Transportation Systems}, vol.~25, no.~6, pp. 5743--5754, 2024.

\bibitem{dong2022federated}
J.~Dong, L.~Wang, Z.~Fang, G.~Sun, S.~Xu, X.~Wang, and Q.~Zhu, ``Federated class-incremental learning,'' in \emph{IEEE/CVF Conference on Computer Vision and Pattern Recognition (CVPR)}, June 2022.

\bibitem{10529194}
Q.~Lin, Y.~Li, W.-B. Kou, T.-H. Chang, and Y.-C. Wu, ``Communication-efficient activity detection for cell-free massive mimo: An augmented model-driven end-to-end learning framework,'' \emph{IEEE Transactions on Wireless Communications}, pp. 1--1, 2024.

\bibitem{lin2023communication}
Q.~Lin, Y.~Li, W.~Kou, T.-H. Chang, and Y.-C. Wu, ``Communication-efficient joint signal compression and activity detection in cell-free massive mimo,'' in \emph{ICC 2023-IEEE International Conference on Communications}.\hskip 1em plus 0.5em minus 0.4em\relax IEEE, 2023, pp. 5030--5035.

\bibitem{wang2021addressing}
L.~Wang, S.~Xu, X.~Wang, and Q.~Zhu, ``Addressing class imbalance in federated learning,'' in \emph{Proceedings of the AAAI Conference on Artificial Intelligence}, vol.~35, no.~11, 2021, pp. 10\,165--10\,173.

\bibitem{huang2021personalized}
Y.~Huang, L.~Chu, Z.~Zhou, L.~Wang, J.~Liu, J.~Pei, and Y.~Zhang, ``Personalized cross-silo federated learning on non-iid data,'' 2021.

\bibitem{li2020federated}
T.~Li, A.~K. Sahu, M.~Zaheer, M.~Sanjabi, A.~Talwalkar, and V.~Smith, ``Federated optimization in heterogeneous networks,'' 2020.

\bibitem{acar2021federated}
D.~A.~E. Acar, Y.~Zhao, R.~Matas, M.~Mattina, P.~Whatmough, and V.~Saligrama, ``Federated learning based on dynamic regularization,'' in \emph{International Conference on Learning Representations}, 2021.

\bibitem{kou2024adverse}
W.-B. Kou, G.~Zhu, R.~Ye, S.~Wang, Q.~Lin, M.~Tang, and Y.-C. Wu, ``An adverse weather-immune scheme with unfolded regularization and foundation model knowledge distillation for street scene understanding,'' \emph{arXiv preprint arXiv:2409.14737}, 2024.

\bibitem{kou2025label}
W.-B. Kou, G.~Zhu, R.~Ye, S.~Wang, M.~Tang, and Y.-C. Wu, ``Label anything: An interpretable, high-fidelity and prompt-free annotator,'' \emph{arXiv preprint arXiv:2502.02972}, 2025.

\bibitem{10342102}
Z.~Zhengl, Y.~Chen, B.-S. Hua, and S.-K. Yeung, ``Compuda: Compositional unsupervised domain adaptation for semantic segmentation under adverse conditions,'' in \emph{2023 IEEE/RSJ International Conference on Intelligent Robots and Systems (IROS)}, 2023, pp. 7675--7681.

\bibitem{zhou2022rethinking}
T.~Zhou, W.~Wang, E.~Konukoglu, and L.~Van~Gool, ``Rethinking semantic segmentation: A prototype view,'' in \emph{Proceedings of the IEEE/CVF Conference on Computer Vision and Pattern Recognition}, 2022, pp. 2582--2593.

\bibitem{10341519}
A.~Z. Zhu, J.~Mei, S.~Qiao, H.~Yan, Y.~Zhu, L.-C. Chen, and H.~Kretzschmar, ``Superpixel transformers for efficient semantic segmentation,'' in \emph{2023 IEEE/RSJ International Conference on Intelligent Robots and Systems (IROS)}, 2023, pp. 7651--7658.

\bibitem{9697426}
Y.~B. Can, A.~Liniger, O.~Unal, D.~Paudel, and L.~Van~Gool, ``Understanding bird’s-eye view of road semantics using an onboard camera,'' \emph{IEEE Robotics and Automation Letters}, vol.~7, no.~2, pp. 3302--3309, 2022.

\bibitem{chen2018encoderdecoder}
L.-C. Chen, Y.~Zhu, G.~Papandreou, F.~Schroff, and H.~Adam, ``Encoder-decoder with atrous separable convolution for semantic image segmentation,'' 2018.

\bibitem{zhang2022topformer}
W.~Zhang, Z.~Huang, G.~Luo, T.~Chen, X.~Wang, W.~Liu, G.~Yu, and C.~Shen, ``Topformer: Token pyramid transformer for mobile semantic segmentation,'' in \emph{Proceedings of the IEEE/CVF Conference on Computer Vision and Pattern Recognition}, 2022, pp. 12\,083--12\,093.

\bibitem{hsu2020federated}
T.~M.~H. Hsu, H.~Qi, and M.~Brown, ``Federated visual classification with real-world data distribution,'' in \emph{Computer Vision--ECCV 2020: 16th European Conference, Glasgow, UK, August 23--28, 2020, Proceedings, Part X 16}.\hskip 1em plus 0.5em minus 0.4em\relax Springer, 2020, pp. 76--92.

\bibitem{li2021model}
Q.~Li, B.~He, and D.~Song, ``Model-contrastive federated learning,'' in \emph{Proceedings of the IEEE/CVF conference on computer vision and pattern recognition}, 2021, pp. 10\,713--10\,722.

\end{thebibliography}
\end{document}